\documentclass[10pt,twocolumn]{article}

\usepackage[T1]{fontenc}
\usepackage[utf8]{inputenc}
\usepackage{lmodern}
\usepackage{microtype}
\usepackage{amsmath}
\usepackage{amssymb}
\usepackage{booktabs}
\usepackage{tabularx}
\usepackage{multirow}
\usepackage{array}
\usepackage{graphicx}
\usepackage{xcolor}
\usepackage[font=small,labelfont=bf,skip=4pt]{caption}
\usepackage{textcomp}
\usepackage{xspace}
\usepackage{url}
\usepackage[numbers,sort&compress]{natbib}

\usepackage[raggedright]{titlesec}

\usepackage[margin=1.9cm,top=2.1cm,bottom=2.2cm]{geometry}
\usepackage[hidelinks,breaklinks=true]{hyperref}
\usepackage{orcidlink}
\usepackage[nameinlink,capitalise]{cleveref}

\hypersetup{
  pdftitle={Most of the LLM Routing Gap Is Task Type: A Task-Type and Language
            Decomposition over a Fully Executed 14 x 294 Matrix, and the
            Reproducibility Floor It Has to Clear},
  pdfauthor={Janghoon Lee},
  pdfsubject={LLM routing; model selection; cost-quality trade-off;
              multilingual evaluation; reproducibility},
  pdfkeywords={LLM routing, model selection, router benchmark,
               reproducibility, nondeterministic inference, multilingual
               evaluation, Korean, Hindi, cost-quality tradeoff}}
\crefformat{section}{\S#2#1#3}
\crefformat{subsection}{\S#2#1#3}
\Crefformat{section}{\S#2#1#3}
\Crefformat{subsection}{\S#2#1#3}
\crefrangeformat{section}{\S#3#1#4--\S#5#2#6}
\crefmultiformat{section}{\S#2#1#3}{ and~\S#2#1#3}{, \S#2#1#3}{ and~\S#2#1#3}
\crefmultiformat{subsection}{\S#2#1#3}{ and~\S#2#1#3}{, \S#2#1#3}{ and~\S#2#1#3}

\usepackage{pgfplots}
\pgfplotsset{compat=1.18}
\usetikzlibrary{patterns}

\newcolumntype{R}{>{\raggedleft\arraybackslash}X}
\newcolumntype{L}{>{\raggedright\arraybackslash}X}
\newcommand{\hd}[1]{\textbf{\small #1}}


\newcommand{\Nmodels}{14\xspace}             
\newcommand{\Nitems}{294\xspace}             
\newcommand{\Npercell}{14\xspace}            
\newcommand{\Ntlcells}{21\xspace}            
\newcommand{\Ncells}{4,116\xspace}           
\newcommand{\Nkeys}{8,232\xspace}            
\newcommand{\Nscoredkeys}{8,229\xspace}      
\newcommand{\Nledger}{98\xspace}             
\newcommand{\Ledgerhash}{903af87edfaa\xspace} 

\newcommand{\Nflips}{221\xspace}             
\newcommand{\Fliprate}{5.37\%\xspace}        
\newcommand{\Ncomparable}{4,114\xspace}      
\newcommand{\BilledA}{\$51.557488\xspace}    
\newcommand{\BilledB}{\$49.591932\xspace}    
\newcommand{\Billeddelta}{$-$3.81\%\xspace}  
\newcommand{\Billedmove}{3.81\%\xspace}      
\newcommand{\Smokeitems}{9\xspace}           
\newcommand{\Smokerate}{3.06\%\xspace}       

\newcommand{\OracleStableAll}{274\xspace}    
\newcommand{\BestStableAll}{245\xspace}      
\newcommand{\GapStableAll}{29\xspace}        
\newcommand{\OracleNaiveA}{280\xspace}       
\newcommand{\BestNaiveA}{249\xspace}         
\newcommand{\OracleNaiveB}{277\xspace}       
\newcommand{\BestNaiveB}{252\xspace}         

\newcommand{\BetweenNaive}{19\xspace}        
\newcommand{\GapNaive}{31\xspace}            
\newcommand{\ShareNaive}{61.3\%\xspace}      
\newcommand{\BetweenStable}{21\xspace}       
\newcommand{\WithinStable}{8\xspace}         
\newcommand{\ShareStable}{72.4\%\xspace}     
\newcommand{\OtherSixNaive}{3\xspace}        
\newcommand{\OtherSixNaiveRate}{1.02\%\xspace} 

\newcommand{\LangOnlyNaive}{254\xspace}      
\newcommand{\LangOnlyStable}{248\xspace}     
\newcommand{\TaskOnlyStable}{266\xspace}     
\newcommand{\ArgmaxNaive}{271\xspace}        
\newcommand{\ArgmaxStable}{268\xspace}       
\newcommand{\StructureStable}{23\xspace}     
\newcommand{\LangGainArgmaxStable}{2\xspace} 
\newcommand{\ResidualArgmax}{6\xspace}       
\newcommand{\ResidualArgmaxRate}{2.04\%\xspace} 
\newcommand{\ResidualArgmaxNaive}{9\xspace}  
\newcommand{\ResidualArgmaxNaiveRate}{3.06\%\xspace} 
\newcommand{\ResidualTaskOnlyNaive}{12\xspace} 
\newcommand{\ResidualTaskOnlyStable}{8\xspace} 

\newcommand{\LookupCorrect}{262\xspace}      
\newcommand{\LookupCost}{\$3.332910\xspace}  
\newcommand{\LookupCostShort}{\$3.33\xspace} 
\newcommand{\LookupCostA}{\$3.598971\xspace} 
\newcommand{\LookupCostB}{\$3.066849\xspace} 

\newcommand{\OpusCost}{\$7.688898\xspace}    
\newcommand{\OpusCostShort}{\$7.69\xspace}   
\newcommand{\OpusCostB}{\$7.066535\xspace}   
\newcommand{\LookupVsOpusItems}{17\xspace}   
\newcommand{\LookupVsOpusCost}{\$4.355988\xspace} 

\newcommand{\UniqueWinnerCorrect}{267\xspace} 
\newcommand{\UniqueWinnerCost}{\$3.368173\xspace} 
\newcommand{\WidePoolCorrect}{261\xspace}    
\newcommand{\WidePoolCost}{\$3.326753\xspace} 
\newcommand{\UniformLoss}{5\xspace}          
\newcommand{\UniformLossRate}{1.70\%\xspace} 

\newcommand{\FrozenCorrect}{264\xspace}      
\newcommand{\FrozenCost}{\$0.542932\xspace}  
\newcommand{\SymmetricCorrect}{259\xspace}   
\newcommand{\SymmetricCost}{\$0.507669\xspace} 
\newcommand{\LookupVsSymItems}{3\xspace}     
\newcommand{\LookupVsSymRate}{1.02\%\xspace} 
\newcommand{\LookupVsSymCost}{\$2.825241\xspace} 
\newcommand{\LookupVsSymCostShare}{84.8\%\xspace} 
\newcommand{\LookupVsFrozenCostShare}{83.7\%\xspace} 

\newcommand{\Nvariants}{7\xspace}            
\newcommand{\TopTwoMargin}{8\xspace}         
\newcommand{\TopTwoRate}{2.72\%\xspace}      

\newcommand{\LookupAcc}{89.12\%\xspace}      
\newcommand{\Nitemscapped}{293\xspace}       
\newcommand{\LookupSpread}{$-$14.79\%\xspace} 
\newcommand{\CodingKoSpread}{$-$17.0\%\xspace} 
\newcommand{\CodingCost}{\$3.107373\xspace}  
\newcommand{\CodingKoCost}{\$2.924815\xspace} 

\newcommand{\CodingCostShare}{93.2\%\xspace} 
\newcommand{\CodingKoShare}{87.8\%\xspace}   

\newcommand{\VariantMinPts}{7.3\xspace}      
\newcommand{\VariantMaxPts}{10.5\xspace}     
\newcommand{\VariantMidPts}{8.5\xspace}      
\newcommand{\VariantNaiveBPts}{9.9\xspace}   
\newcommand{\SolCorrect}{239\xspace}         
\newcommand{\LunaCorrect}{223\xspace}        

\newcommand{\GateDistancePts}{9.86\xspace}   

\title{\bfseries Most of the LLM Routing Gap Is Task Type\\[2pt]
\large\mdseries A Task-Type and Language Decomposition over a Fully Executed
$14\times294$ Matrix,\\ and the Reproducibility Floor It Has to Clear}

\author{Janghoon Lee\,\orcidlink{0009-0002-8108-5407} \\
        Redrob \\
        \texttt{janghoon@redrob.io} \\
        {\small ORCID 0009-0002-8108-5407}}
\date{}

\begin{document}

\twocolumn[
  \begin{@twocolumnfalse}
    \maketitle
    {\setlength{\parskip}{5pt}\section*{Abstract}

An LLM router picks which model should answer each query. The appeal is that models fail on different questions. Whatever single model is best overall still gets some wrong, and another model in the pool gets many of those right. Route each query to the model that will handle it and you beat any one model at lower cost. Getting that choice right every time is the ceiling, and a router is an attempt to approach it.

However, recent work reports that routers do not get close. Across 21 routing methods on five benchmarks, sharply different designs land within a fraction of a point of each other, and all of them stay far below that ceiling. Learned routers often fail to beat simply always calling the strongest model. Something is limiting them, and it is not the router design.

We ask what those missed questions have in common. We set fourteen models to answer all \Nitems questions, with 7 task types across 3 languages: Korean, English and Hindi. We ran the whole matrix twice, changing nothing, but \Fliprate of the \Ncells model-question pairs came out scored differently anyway. Run-to-run movement like that is normal, and we argue that a small win does not show that routing did anything, ours or anyone else's.

Counting an answer correct only when the model got it right in both runs, 29 questions on this matrix can be improved with routing. The best single model gets them wrong and some other model gets them right. Every correct-answer count in this paragraph is on that rule. Task type accounts for most of them. Assigning each task type one model in advance, chosen once and never updated, improves \BetweenStable of the \GapStableAll. Splitting each task type by language improves 2 more and leaves \ResidualArgmax of \Nitems unoptimized. That handful is what a learned router would have been built for. It is smaller than the run-to-run movement mentioned above, which is a share of model-question pairs rather than of questions. The static table we adopted answers \LookupCorrect of \Nitems questions at \LookupCostShort per run, against the best single model's \BestStableAll at \OpusCostShort.

All of this is fitted and scored on the same \Nitems questions with no holdout.
}
    \vspace{6mm}
  \end{@twocolumnfalse}
]

\section{Introduction}\label{sec:intro}

No single model is best at everything. Models complement each other in accuracy and in cost, and none of them is universally dominant \citep{routerbench2024,llmrouterbench2026}. That is the premise of LLM routing. Combining models under a per-query choice should overcome the constraints of any individual one \citep{routerbench2024}, at less than the cost of always calling the strongest \citep{routellm2024}. The ceiling is the oracle router, which picks in hindsight a model that answers each query correctly, breaking ties by cost. The floor is the best single model, the one model in the pool with the highest score over the whole workload. The question is how much of the distance between the two a real router recovers.

Recent work answers: a similar fraction, whatever the router. Across 21 routing methods on five benchmarks under one setup, \citet{plateau2026} report a \emph{routing plateau}. Sharply different designs converge into a narrow band: the top five are separated by 0.22 percentage points (pp) on average, and the top fifteen on RouterBench by 0.23 pp. The best still trails the oracle by 10 to 30 pp. A nearest-neighbour router over a frozen text encoder ranks in the top two on all five. Re-evaluating 33 models on 21 datasets, \citet{llmrouterbench2026} find the same convergence. They report that several recent routers, including commercial ones, do not reliably beat a simple best-single-model baseline. They attribute the surviving oracle gap primarily to model-recall failure: routers not selecting the candidate that would have been correct. This paper is downstream of those findings.

What that does not settle is what the gap is made of, in coordinates a policy could condition on. Existing decompositions cut it along other axes. Query difficulty is one: the hardest 11--35\% of queries carry 70--91\% of the gap \citep{plateau2026}. Dataset identity is another. There, a small distance to a dataset-level oracle is read as an indication that much of the available gain may be coarse-grained domain structure \citep{llmrouterbench2026}. The third is label stochasticity, separating reproducible specialist advantage from single-draw noise \citep{routinggapnoise2026}. Of those three coordinates, difficulty is not observable at routing time, and label stochasticity is not a routing choice. Dataset identity is observable, and it is the coordinate just named as a candidate for much of the available gain. Language is missing altogether. Multilingual items do appear in the pools. The multilingual question-answering sets MLQA, TyDi QA and XQuAD are among the sources of Nine-by-30k, one of the plateau study's five benchmarks. But we find no routing result broken out by language in RouterBench \citep{routerbench2024}, RouterEval \citep{routereval2025}, LLMRouterBench \citep{llmrouterbench2026}, or the five benchmarks of the plateau study \citep{plateau2026}. RouterBench lists non-English long-tail tasks as future coverage. Whether the gap has the same size and composition in Korean and Hindi as in English is therefore not answered there. That is why this study carries a language axis.

We take a small matrix and execute all of it. Fourteen models, seven task types, three languages, 14 items per task$\times$language cell: \Nitems queries per model and \Ncells model-query cells. We executed the matrix twice at temperature 0 and seed 42 through one gateway, for \Nkeys keys, of which \Nscoredkeys are scored (\Cref{sec:setup}). A key is one model-query cell in one named run. The oracle, the best single model and any static policy are readable off the same frozen matrix, since every model answers every query. No router has to be trained to ask where the gap sits. Two scoring rules appear below. The single-run rule reads one run. The main one, the both-run rule, scores an item correct for a model only when that model is correct in both (\Cref{sec:decomp:2}). n per task$\times$language cell is 14, so no claim here rests on a single task$\times$language cell.

First, we decompose the oracle-minus-best-single gap by task type and language. Under the single-run rule read on the first run, the \GapNaive-item gap splits into \BetweenNaive items a per-task-type choice recovers and 12 left inside a single task type. A lookup on task type alone therefore accounts for \BetweenNaive/\GapNaive = \ShareNaive of it, with no learned component. That lookup is fitted by taking the best model in each task type on this same matrix. Under the both-run rule the split is \BetweenStable of \GapStableAll, \ShareStable. Adding language leaves \ResidualArgmax of the \Nitems items (\ResidualArgmaxRate) under the both-run rule and 9 (\Smokerate) under the single-run one. These are in-sample shares, fitted and evaluated on the same \Nitems items with no holdout (\Cref{sec:limits}). The remainder sits in one task type. Of the 8 items left inside a task type under the both-run rule, 7 are coding, and language recovers 2 of those 8. The reading this supports is narrow. Here the residual after conditioning on task type and language, \ResidualArgmaxRate of items, is below both of the reference magnitudes the next paragraph reports. The first is the \Fliprate of model-query cells whose correct bit changes when we execute the identical configuration again. The second is a per-item reference that is numerically the same \Smokerate as the single-run residual just given, and a different quantity from it. The first of those two rates is carried on a different denominator from the residual, and \Cref{sec:noise:2} states the procedure under which both comparisons are made.

Second, we report that movement, in the body rather than an appendix, because \Cref{sec:decomp} and \Cref{sec:rule} are unreadable without it. We ran the identical \Ncells-model-query-cell matrix twice at temperature 0 and seed 42 through the same gateway. Between those runs, \Nflips scored correct bits flip (\Fliprate) and billed cost moves \Billeddelta, from \BilledA to \BilledB. An earlier check gives a third reference of at least 9/\Nitems = \Smokerate of items (\Cref{sec:noise}). In that check one model answered the same \Nitems items before and after an output-cap change, rather than under an identical configuration. Its \Smokerate is numerically the same fraction as the single-run residual above, and a different quantity. We treat an accuracy advantage at or below either of those two magnitudes as unresolved by this setup, ours included, by the procedure \Cref{sec:noise:2} states. The two magnitudes are \Smokerate of items and the \Fliprate of model-query cells, which is a rate on a different denominator. None of the three references this paragraph names, the billed movement included, was fixed in advance. This is not the sampling noise of an oracle assembled from single stochastic draws \citep{routinggapnoise2026}, estimated there by resampling at non-zero temperature. It is the residual irreproducibility of a nominally deterministic configuration.

Third, on this matrix which model is best is a property of the scoring rule rather than of the matrix. Five different models hold the title across seven scoring variants and the three per-language strata of the main variant. No top-two margin exceeds \TopTwoRate in accuracy. A scoring variant is a scoring rule paired with a rule for which items count. Six of the ten compare two models on one denominator. The largest of those six, 8/\Nitems = \TopTwoRate, is below the \Smokerate per-item reference and well below the \Fliprate model-query-cell flip rate. The other four compare ratios on per-model denominators and are reported as sizes (\Cref{sec:rule:1}). The winner of the main variant scores \BestNaiveA and \BestNaiveB items on the two runs under the single-run rule. That 3-item move on identical inputs is larger than the single item that decides the main variant. The best single model is one of the reference baselines against which routing gains are quoted. Gain@B, the accuracy gain at a cost budget B, is measured against it \citep{llmrouterbench2026}. On this matrix that baseline moves with the scoring rule alone, before any router is involved. We report all seven variants rather than one.

None of this shows that routers are unnecessary. The scope is 14 models from one catalog snapshot, 7 task types, 3 languages, 14 items per task$\times$language cell, one gateway, two runs, and an in-sample policy comparison. What we defend is the narrow statement above. It points the same way as the plateau results, from the other side. They measured that routers converge below the oracle. We locate one part of where the distance goes.

\section{Setup}\label{sec:setup}

This paper reports a pilot: one measurement campaign, designed and configured in advance, run to completion, and then closed. Its design is a full matrix. Seven task types $\times$ 3 languages $\times$ 14 items give \Nitems queries, every one of the 14 models answers all \Nitems, and one run is therefore \Ncells model-query cells. The entire matrix was executed twice, giving \Nkeys keys, one per model-query cell per run. The replicate exists because temperature 0 and a fixed seed do not make this stack deterministic; \Cref{sec:noise} quantifies how much it moves. Counts are last-per-key, so infrastructure retries are not double-counted and no separate raw HTTP call total is reported. All calls use temperature 0 and seed 42 through one gateway (OpenRouter). Both runs are complete at \Ncells/\Ncells keys. Three of those keys remained infrastructure failures after the retry budget and carry no score, so \Nscoredkeys of the \Nkeys keys are scored. Those three are two keys in the first run and one in the second. All are on the same Korean coding item, and they cover two of the \Ncells positions.

Four counts recur, and the paper keeps a separate word for each. An \textbf{item} is one of the \Nitems queries a model sees; accuracy figures are counts of items on a denominator of \Nitems unless stated otherwise. A \textbf{model-query cell} is one position in the matrix, one model paired with one item, of which there are \Ncells; each run fills all of them. A \textbf{key} names a model-query cell together with the run it was filled in, so the two runs give \Nkeys keys. A quantity counted per key is counted once per execution rather than once per position. A \textbf{cell}, unqualified, is one of the 21 task$\times$language cells the lookup of \Cref{sec:lookup} is keyed on, each holding 14 items. The abstract calls an item a question, which is what it is; the body says item throughout.

Per-cell n is 14. That is enough for the whole-matrix comparisons in \Cref{sec:decomp} and \Cref{sec:lookup} and too small to support a claim about any one cell. Task$\times$language numbers are meant to be read as components of an aggregate. \Cref{sec:limits:3} records the places where the paper does not hold to that and what each of them costs.

\subsection{Model pool}\label{sec:setup:1}

Prices are USD per million tokens, from the \mbox{2026-08-21} catalog snapshot, and are configuration only: the cost axis is billed cost (\Cref{sec:setup:5}). Model identifiers were verified by live call on \mbox{2026-08-22}, with 14/14 present in the gateway catalog and the routed model equal to the requested model in every case.

\begin{table*}[t]
\centering\footnotesize
\setlength{\tabcolsep}{5pt}
\caption{The 14-model pool. The short names are the identifiers the configuration and the artifacts use; prose names each model in full, and the other tables in this paper key on the short names. Prices are USD per million tokens from the \mbox{2026-08-21} catalog snapshot and are configuration only; the cost axis is billed cost (\Cref{sec:setup:5}).}
\label{tab:s2-setup-1}
\begin{tabularx}{\textwidth}{@{}>{\hsize=0.754\hsize\raggedright\arraybackslash}Xl>{\hsize=1.246\hsize\raggedright\arraybackslash}Xrr@{}}
\toprule
\hd{short name} & \hd{vendor} & \hd{model id} & \hd{in} & \hd{out} \\
\midrule
\texttt{gpt\_5\_6\_sol} & OpenAI & \texttt{openai/gpt-5.6-sol} & 5.00 & 30.00 \\
\texttt{gpt\_5\_6\_terra} & OpenAI & \texttt{openai/gpt-5.6-terra} & 2.00 & 12.00 \\
\texttt{gpt\_5\_6\_luna} & OpenAI & \texttt{openai/gpt-5.6-luna} & 0.20 & 1.20 \\
\texttt{claude\_fable\_5} & Anthropic & \texttt{anthropic/claude-fable-5} & 10.00 & 50.00 \\
\texttt{claude\_opus\_5} & Anthropic & \texttt{anthropic/claude-opus-5} & 5.00 & 25.00 \\
\texttt{claude\_sonnet\_5} & Anthropic & \texttt{anthropic/claude-sonnet-5} & 2.00 & 10.00 \\
\texttt{gemini\_3\_1\_pro} & Google & \texttt{google/gemini-3.1-pro-preview} & 2.00 & 12.00 \\
\texttt{gemini\_3\_7\_flash} & Google & \texttt{google/gemini-3.7-flash} & 0.75 & 3.75 \\
\texttt{grok\_4\_6} & xAI & \texttt{x-ai/grok-4.6} & 2.00 & 6.00 \\
\texttt{grok\_4\_3} & xAI & \texttt{x-ai/grok-4.3} & 1.25 & 2.50 \\
\texttt{deepseek\_v4\_pro} & DeepSeek & \texttt{deepseek/deepseek-v4-pro} & 0.435 & 0.87 \\
\texttt{deepseek\_v4\_flash} & DeepSeek & \texttt{deepseek/deepseek-v4-flash} & 0.14 & 0.28 \\
\texttt{kimi\_k3} & Moonshot & \texttt{moonshotai/kimi-k3} & 2.60 & 13.00 \\
\texttt{qwen3\_7\_plus} & Alibaba & \texttt{qwen/qwen3.7-plus} & 0.32 & 1.28 \\
\bottomrule
\end{tabularx}
\end{table*}

The pool of Table 1 pairs a higher and a lower tier from the same vendor wherever both exist. The assumption is that the tier spread inside a vendor is larger than the spread between vendors at a fixed tier. Routing that never crosses tiers would have little to choose from.

Eight of the fourteen list prices carry conditions recorded in the configuration. They are introductory rates with known expiry, long-context meters above 200K tokens, gateway list prices that differ from the vendor's direct prices, and input prices cut shortly before the snapshot. None of them enters a reported accuracy or cost figure, for the reason given in \Cref{sec:setup:5}. \Cref{sec:limits:7} enumerates all eight and names the one place two of them are used. That place is a reconstruction on an earlier one-model trial run, reported as a disagreement between price tables and not as a cost of anything.

\subsection{Task types and sources}\label{sec:setup:2}

The seven task types are the seven rows of Table 2, called \texttt{bucket} in the configuration and in the artifacts. Each draws every one of its items from a single source. Task type and source are therefore the same partition on this matrix, which is what \Cref{sec:limits:2} is about. Revisions are dataset commit or revision hashes as of \mbox{2026-08-21}. The ``translated field'' column is the only text sent to the translator; \Cref{sec:setup:3} gives the procedure.

\begin{table*}[t]
\centering\footnotesize
\setlength{\tabcolsep}{5pt}
\caption{The seven task types, their sources, and how each is scored. One source per task type, which is the confound \Cref{sec:limits:2} states.}
\label{tab:s2-setup-2}
\begin{tabularx}{\textwidth}{@{}l>{\hsize=1.351\hsize\raggedright\arraybackslash}Xlll>{\hsize=0.649\hsize\raggedright\arraybackslash}X@{}}
\toprule
\hd{task type} & \hd{source} & \hd{revision} & \hd{license} & \hd{scorer} & \hd{translated field} \\
\midrule
knowledge & MMLU-ProX (\texttt{li-lab/MMLU-ProX}) & \texttt{8e6106a6\dots{}} & MIT & \texttt{mcq} & none (parallel source) \\
math & MGSM (\texttt{juletxara/mgsm}, config \texttt{en}) & \texttt{b2f13d42\dots{}} & CC BY-SA 4.0 & \texttt{exact} & statement \\
instruction & IFEval (\texttt{google/IFEval}) & \texttt{966cd895\dots{}} & Apache-2.0 & \texttt{ifeval} & prompt (instruction arguments frozen) \\
extraction & in-house & n/a & Apache-2.0 & \texttt{json\_match} & document body \\
toolcall & BFCL v4 (\texttt{ShishirPatil/gorilla}) & \texttt{6ea57973\dots{}} & Apache-2.0 & \texttt{ast} & utterance \\
abstention & BFCL v4 irrelevance (same repo) & \texttt{6ea57973\dots{}} & Apache-2.0 & \texttt{abstain} & utterance \\
coding & LiveCodeBench (\texttt{livecodebench/code\_generation\_lite}, \texttt{release\_v6}) & \texttt{0fe84c39\dots{}} & MIT (harness) & \texttt{exec} & statement head \\
\bottomrule
\end{tabularx}
\end{table*}

Sampling within each task type is seeded and stratified where the source supports it: BFCL by call type (4/4/3/3), LiveCodeBench by difficulty (3/5/6), IFEval by instruction type. IFEval takes 2 items each over 7 adopted types. They are drawn from the 86 items whose types do not grade against one of the instruction's own arguments, which translating would have altered. IFEval records those arguments as \texttt{kwargs}. Math items carry the required-quantity annotation joined from GSM8K (\texttt{740312ad\dots{}}, MIT) by exact English question match; the join is a translation-checking pin, not grading data. Extraction is the one in-house task type: 14 short documents over nine patterns, each with a JSON gold record whose keys and categorical values are English. The patterns are invoice, job posting, appointment, event, and product twice each, and shipment, hotel, licence, and sensor once. Scoring is exact key-value match, and extra keys fail. It is not part of any public release. Abstention is scored as correct when the model makes no tool call at all; any parsed call is incorrect.\footnote{A translation task type was considered and dropped. In a design that crosses task with query language, translation is the one task whose content \emph{is} the language pair, so its language condition is not a free variable.}

The LiveCodeBench window is \mbox{2025-01-01} to \mbox{2025-04-06}, the end of the release, and there is no contamination control: the configuration records that these items are not post-cutoff data for most of the pool. \Cref{sec:limits} returns to what that does to the coding task type.

\subsection{Language conditions}\label{sec:setup:3}

The three conditions are Korean, English, and Hindi. The design invariant is that one item identifier denotes the same item in all three: 98 unique item identifiers $\times$ 3 languages = \Nitems queries. The frozen 98-item ledger is fixed under SHA-256 \texttt{903af87e\dots{}}. Neither the ledger nor the label matrix accompanies this version, so that hash is a commitment and not yet something a reader can check. It fixes which 98 identifiers every number in this paper was computed over, and anything released later can be checked against it. If that invariant broke, language effects and item difficulty would be confounded and the matrix would be unreadable.

Knowledge uses the existing MMLU-ProX ko/en/hi parallel release and is the only task type we did not translate. The other six were translated from English by a single pipeline. That pipeline is \texttt{openai/gpt-4.1}, temperature 0, seed 42, the same template for both target languages, and a per-task-type field split that sends only the field named in Table 2. Everything else is held in English by construction: numbers, formulas, code, tests, identifiers, function schemas, JSON keys, proper names, dates, and instruction markers and arguments. For toolcall and abstention this is enforced at the interface: the function schema is never sent to the translator, only the user utterance. The categorical values a gold record can take are the same English strings in all three languages. Those values are the extraction schema's enumerated values and the abstention gold \texttt{no\_tool}.

Item replacement during construction is capped at five per task type, and unreplaced items reduce n rather than being topped up from the pool. Resampling until an item passes is not allowed. Extraction has no replacement pool at all, being constructed rather than sampled. Translations were validated before the model runs; the procedure, the realized replacement counts, and the validation results are in a companion note (in preparation) \citep{companion}.

\subsection{Scoring and status}\label{sec:setup:4}

Every scored model-query cell is 0 or 1 under the task type's scorer. The aggregation rules were pinned on \mbox{2026-08-21} and not changed afterwards. Output caps are a separate pin. The values below were raised the next day to correct a configuration error, after the aggregation pin and before any of the 14-model runs reported here.\footnote{The aggregation pin is \mbox{2026-08-21T13:22Z}, committed a minute later at \texttt{3be49a0}; the cap correction is \texttt{e5932f3}, \mbox{2026-08-22T07:26:52Z}. Both are recorded in the configuration and in the run reports.} Every model-query cell carries a status, and the status determines the denominator:

\begin{itemize}\setlength{\itemsep}{1pt}
\item \texttt{ok}: completed; the scorer decides.
\item \texttt{parse\_failed}: completed with no usable answer, including empty content; \texttt{correct = 0} and the item stays in the \Nitems denominator.
\item \texttt{cap\_hit}: the output cap was reached. The pinned detection rule is \texttt{finish\_reason in [length, max\_tokens] or completion\_tokens >= cap}, the two finish reasons being gateway aliases for one event. \texttt{correct = 0}, also in the denominator, and counted separately from \texttt{parse\_failed}. Caps are 8,192 tokens for every task type except coding, which is 32,768.
\item \texttt{sandbox\_error}, \texttt{http\_error}, \texttt{timeout}: infrastructure, not model behaviour. These are retried up to three times; if they still fail the model-query cell is left null and excluded, including from the oracle.
\end{itemize}

The primary denominator is the full \Nitems-item matrix with \texttt{parse\_failed} and \texttt{cap\_hit} at 0. An infrastructure null reduces that model's denominator by one, and where it occurs both denominators are shown. Two diagnostic denominators exist, parse-ok items only and the \texttt{cap\_hit} split. \Cref{sec:decomp} reports both as sensitivity variants rather than substituting either for the main figure. The two are kept apart because a cap is our design decision and a parse failure is the model missing the output contract.

Model outputs are not stored. Each run writes a label row of item, model, status, correct bit, token counts, cost and latency. The writer rejects any row containing a response field, which is checked per run.

\subsection{Cost}\label{sec:setup:5}

The cost axis is billed \texttt{usage.cost} as reported by the gateway for each request, pinned before any model scores were seen. Two reconstructions are possible from the same token counts: the vendor catalog prices in the configuration, and the gateway's list prices. Neither agrees with the billed amount or with the other. We record that disagreement and do not reconstruct. Billed and catalog figures are never mixed within a model, and no reported cost is derived from a price table.

When a request returns no billed cost, the item is excluded from that model's cost total while the model keeps its accuracy; no catalog value is imputed. \Cref{sec:lookup} states where this applies.

\section{Execution noise floor}\label{sec:noise}

Executing the same matrix twice under the same configuration does not return the same scores. This section measures how much moves (\Cref{sec:noise:1}) and then fixes the procedure by which the rest of the paper reads a difference against that movement (\Cref{sec:noise:2}).

\subsection{What re-execution moves}\label{sec:noise:1}

The scored-correct bit moves between two executions of the same matrix: \Nflips of \Ncells model-query cells flip between A and B (\Fliprate). The matrix is 14 models $\times$ \Nitems items = \Ncells model-query cells per run. Run A is the first execution of that matrix and run B the second; both use the same queries, temperature 0, and seed 42. Of those flips, 115 are 1$\rightarrow$0 and 106 are 0$\rightarrow$1. The rate is quoted over all \Ncells model-query cells. Two of them are unscored in at least one run, since the three infrastructure failures of \Cref{sec:setup} fall on two positions. So \Ncomparable are actually comparable, and the rate is \Fliprate under either denominator. Temperature 0 and seed 42 are not bit-identical on this stack.

The cost axis has its own floor. The billed cost of the whole matrix is \BilledA on A and \BilledB on B, a \Billeddelta move. That figure is the cost-axis reference for later billed comparisons. It is not the same estimand as the \Nflips/\Ncells correct-bit flips.

A third reference predates both runs and is weaker than either. Before the matrix was executed, one model in the pool, DeepSeek V4 Flash, answered all \Nitems items twice. The first of those runs used a smaller output-cap configuration than \Cref{sec:setup} states, and the second came after the caps were raised. At least \Smokeitems of the \Nitems items (\Smokerate) that the capped run scored correct were incorrect on the re-run. That comparison changes a configuration as well as re-executing it. It is not a replication of the kind the A/B pair is, which is the weakness \Cref{sec:limits:9} records against it. It is also on a different scale from the flip rate, and the two are kept apart throughout. The \Smokeitems of \Nitems items is a per-item rate for one model, while the \Nflips of \Ncells is a correct-bit flip rate over the full matrix. None of the three was fixed in advance of the results; \Cref{sec:noise:2} states how they are used and what that costs.

The \Nflips flips are not uniform. By model, out of that model's \Nitems model-query cells and split 1$\rightarrow$0 / 0$\rightarrow$1: Claude Sonnet 5 16/12, DeepSeek V4 Flash 16/16, Grok 4.3 11/16, Grok 4.6 7/16, Claude Fable 5 11/6, GPT-5.6 Luna 10/7, Kimi K3 11/4, Qwen3.7 Plus 10/5, DeepSeek V4 Pro 9/9, Claude Opus 5 4/7, GPT-5.6 Terra 4/3, GPT-5.6 Sol 3/4, Gemini 3.1 Pro 3/1, Gemini 3.7 Flash 0/0. The same \Nflips flips split by task type as coding 73, knowledge 49, toolcall 41, extraction 21, abstention 20, instruction 16, math 1. They split by language as Hindi 83, English 74, Korean 64. Gemini 3.7 Flash has zero correct-bit flips. That is a property of this model on this matrix, not evidence that A was copied into B; label rows, latency sums and billed totals still differ.

\subsection{How a difference is read against these references}\label{sec:noise:2}

The rest of the paper reads differences against these references by one stated procedure rather than by a judgement taken again at each site. The procedure is set out here in full, and later sections apply it by name instead of restating it. Two words carry its outcome. A difference this setup \textbf{resolves} is one larger than the movement the references below stand for. A difference it leaves \textbf{unresolved} is one that is not.

What is compared: two quantities measured on this matrix. They are two models' scores, two policies' accuracies, or a policy's residual against the oracle. An accuracy difference is counted in items and read as a rate over the item set it is counted on. That set is \Nitems for a whole-matrix comparison, \Nitemscapped where an infrastructure null leaves an item unscored, and 98 for a single-language one. A billed-cost difference is read as a share of what the costlier of the two policies bills over the same workload. Both sides of every comparison are read as magnitudes, since the references are movements and carry a sign the differences do not. Several comparisons below are between rates that do not share a denominator. One case is a per-item rate set against a per-cell one; another is a \Nitems-item rate against the \Nitemscapped-item one an infrastructure null leaves. The site says so wherever that is the case. Such a comparison is read as one rate standing above or below the other and no further. The gap between two rates on unlike denominators is not itself a rate. The paper never takes one. Where \Cref{sec:limits:9} says how far a reading sits from its deciding reference, it does so by naming the estimate of that reference that would change the reading. One kind of difference is outside the procedure altogether. That is a difference between two ratios on per-model exclusion denominators, where the two models are scored on item sets that differ by construction rather than by one named item. Those are reported as sizes and given no verdict (\Cref{sec:decomp:1}, \Cref{sec:rule:1}).

Which references apply: an accuracy difference is read against both accuracy references. The per-item reference, \Smokeitems of \Nitems items = \Smokerate, is on the same scale as the difference. It is also the one reference of the three not founded on a replication. It comes from one model re-run under a changed cap configuration rather than from the A/B pair. The model-query-cell flip rate, \Nflips of \Ncells = \Fliprate, is founded on that pair, but counts a different event on a different denominator. A per-item difference set beside it is therefore the coarser of the two comparisons. Both are stated because neither is a substitute for the other. A billed-cost difference is read against the \Billeddelta whole-matrix movement. Where a policy's cost concentrates in one cell, it is read against the wider band that concentration produces as well, \LookupSpread for the \Cref{sec:lookup} policy (\Cref{sec:limits:7}). That policy bills \CodingKoShare of its total in a single cell. That band is the same A/B pair measured on one policy and not a fourth reference.

The per-item value is the smaller, so the flip rate decides wherever the two disagree, which is between \Smokerate and \Fliprate, and there the procedure returns unresolved. Reading an accuracy difference against both has that one consequence, and the arithmetic produces it rather than either measurement. Two of the differences this paper puts a verdict on sit in that band: \Cref{sec:rule:1}'s eighth and ninth models, at 3.74\% and 4.42\% behind the leader. Both are unresolved on that conservative branch alone. \Cref{sec:rule:1} gives its count of models under each reference separately for that reason. Every other difference the paper puts a verdict on is at or below \Smokerate or above \Fliprate. Two further differences fall in the band and are reported at their sizes without a verdict, so the branch decides nothing for them. They are the residual a task-type-only rule leaves on single-run A, and the adopted policy's residual to the oracle, both 12 of \Nitems items = 4.08\% (\Cref{sec:decomp:3}, \Cref{sec:lookup:3}). The per-item value is also a lower bound, and the same arithmetic bounds what that costs. A true movement inside the band would change no verdict. Only above \Fliprate would the procedure have resolved differences it should have left unresolved.

What follows from the comparison: a difference at or below any reference that applies to it is reported as unresolved by this setup. The paper gives its size and its direction and draws no conclusion from it, whichever side it favours, ours included. A difference above every reference that applies to it is reported as resolved. That means only that it is larger than what re-executing this same configuration moved here. It does not mean the difference would survive a different item set, a different model pool, or a third run.

The procedure is not preregistered, and it is thin. Neither of those two properties can be read apart from it, and one obligation follows from them. Neither the three references nor this rule for using them was fixed in advance of the results, so it is a reading convention adopted with them in view. It is thin because each reference rests on a single pair of executions, two runs rather than two hundred. The values it compares against therefore carry the uncertainty of an estimate drawn from n = 2 executions. \Cref{sec:limits:9} states which of this paper's readings the smallest change in a reference would reverse, and what that reference would have to be for each of them to reverse. The obligation is that sizes and directions are reported everywhere the procedure is applied, not only its verdicts. A reader who prefers a different threshold can then apply one to the same quantities.

The 5-percentage-point gate of \Cref{sec:decomp:1} is a separate instrument and is not this procedure. It was fixed before the matrix was scored rather than after it, it reads no execution measurement, and none of the three references enters it. Its own weakness is that its timing rests on our record rather than on a dated artifact, which \Cref{sec:decomp:1} states. Two other uses of a reference are neither the procedure nor the gate. \Cref{sec:decomp:2} gives the flip finding as one of the two reasons this paper's main estimate scores an item on both runs. \Cref{sec:lookup:1} treats an accuracy edge whose sign does not survive re-execution as noise, on the strength of the same rate. Both are reasons for a rule and not verdicts on a difference.

\section{Gap decomposition}\label{sec:decomp}

This section asks what the distance between the best single model and a hindsight oracle over the same pool is made of. It states the decision rule the pilot committed to and the estimate it was applied to (\Cref{sec:decomp:1}). It states why that estimate scores an item on both runs rather than one (\Cref{sec:decomp:2}), and where the distance sits (\Cref{sec:decomp:3}).

\subsection{The gate the pilot committed to}\label{sec:decomp:1}

Before the matrix was scored, the pilot fixed one decision rule. Build a learned router only if a per-item oracle over the 14 models beats the best single model by at least 5 percentage points of items. The oracle credits an item whenever at least one model's scored answer to it is correct, that is, whenever any model in the pool answers it correctly. It is therefore the hindsight limit of per-item routing on this pool. The distance it opens over the best single model bounds what any router could recover here. Five points was set as the gap below which a router would not be worth building. Two properties of that rule matter for how the rest of the paper reads. It is a project decision rule, fixed before the matrix was scored and not changed afterwards. But it is recorded only in the pilot's own reports. Unlike the scoring rules of \Cref{sec:setup:4} it carries no dated pre-scoring artifact, so the reader has our word for its timing and not a pin. And it reads no noise floor: the threshold is a flat 5 percentage points of items, fixed without reference to any execution measurement. The three references of \Cref{sec:noise} reach the rest of the paper through the procedure of \Cref{sec:noise:2}. One of them reaches \Cref{sec:decomp:2} and \Cref{sec:lookup:1} as a reason for a rule. No route passes through the gate.

The gate passed. Under the main scoring rule the oracle is \OracleStableAll of \Nitems items and the best single model is \BestStableAll of \Nitems, a distance of \GapStableAll items = \GateDistancePts percentage points. All seven scoring variants keep the oracle above the best single model, with distances between \VariantMinPts and \VariantMaxPts points (Table 3). The three variants that apply no exclusion give \VariantMaxPts, \VariantMidPts and \VariantNaiveBPts points. Each is larger than both accuracy references of \Cref{sec:noise}, so \Cref{sec:noise:2} resolves all three. Those references are 9 of \Nitems items, which is 3.06 points on this scale, and the \Fliprate flip rate. The flip rate is a rate on \Ncells cells and not on these \Nitems items. On those three variants every model is scored on all \Nitems items, except the one or two carrying an infrastructure null, which are scored on \Nitemscapped (\Cref{sec:setup}). The other four variants exclude items per model, so their distances are differences between ratios on unequal denominators and are not comparable to a \Nitems-item reference. Which model is best single model changes with the scoring variant; that identity split is \Cref{sec:rule}. The rest of this section asks where the \GapStableAll items sit.

\subsection{Why the main estimate uses the both-run rule}\label{sec:decomp:2}

This paper's main estimate uses the both-run rule, and the artifacts call the two \texttt{stable} and \texttt{naive}. Two scoring rules are available on a matrix executed twice. The \textbf{both-run rule} counts an item correct for a model only when that model is correct in both runs; the \textbf{single-run rule} reads one run and ignores the other. Two facts decide between them. First, \Cref{sec:noise} found that \Nflips of the \Ncells model-query cells change their correct bit between two executions of the identical configuration. A single-run score therefore credits a model with items that a second execution takes away. The both-run rule is the one of the two whose every correct bit survived a re-execution. Second, a single-run rule has no canonical run. The two runs disagree about the leader: a tie between two models on the first, one of them alone on the second. They also disagree about its score, \BestNaiveA against \BestNaiveB of \Nitems items, and about the distance to the oracle, 31 items against 25. Reporting one of them as the headline would be a choice of run.

The both-run rule is not neutral either. It can only lower a score and never raise one, and it discards the information in items a model answers once. \Cref{sec:rule} shows that the identity of the leader is not settled by adopting it. \Cref{sec:rule} also shows that it partly rewards determinism, since a model with no flips loses nothing to it. The run-A single-run reading is reported alongside the main estimate throughout this section, and the run-B reading for the all-items variant. The figures can then be read under more than one scoring rule.

\begin{figure}[t]
\centering
%
\begin{tikzpicture}
\begin{axis}[
  width=0.86\linewidth, height=3.6cm,
  xbar stacked,
  xmin=0, xmax=46,
  xtick={0,5,10,15,20,25,30},
  xlabel={items of \Nitems\ between oracle and best single model},
  xlabel style={font=\scriptsize},
  ytick={0,1}, yticklabels={single-run (A), both-run},
  tick label style={font=\scriptsize},
  bar width=13pt, y=22pt,
  enlarge y limits={abs=0.7},
  axis lines*=left,
  legend style={font=\scriptsize, at={(0.5,-0.62)}, anchor=north,
                legend columns=2, draw=black!30, column sep=6pt},
]
\addplot[fill=black!15, draw=black, postaction={pattern=north east lines}]
  coordinates {(19,0) (21,1)};
\addlegendentry{between task type}
\addplot[fill=black!55, draw=black, postaction={pattern=crosshatch dots}]
  coordinates {(12,0) (8,1)};
\addlegendentry{within task type}

\node[font=\scriptsize, anchor=west] at (axis cs:31,0) {\;31 items};
\node[font=\scriptsize, anchor=west] at (axis cs:29,1) {\;29 items};
\coordinate (betweenA) at (axis cs:9.5,0);
\coordinate (betweenS) at (axis cs:10.5,1);
\end{axis}
\node[font=\scriptsize, fill=white, inner xsep=2pt, inner ysep=1.2pt]
  at (betweenA) {\ShareNaive};
\node[font=\scriptsize, fill=white, inner xsep=2pt, inner ysep=1.2pt]
  at (betweenS) {\ShareStable};
\end{tikzpicture}
\caption{Where the distance between the oracle and the best single model
sits, under each scoring rule. Under either rule most of it is recovered by
giving each task type its own best model, and the remainder needs a
different model inside a task type. The percentage on each bar is the
between-type share of that bar's total, and the count past its end is the
total. The two components are distinguished by fill pattern as well as by
shade.}
\label{fig:decomposition}
\end{figure}
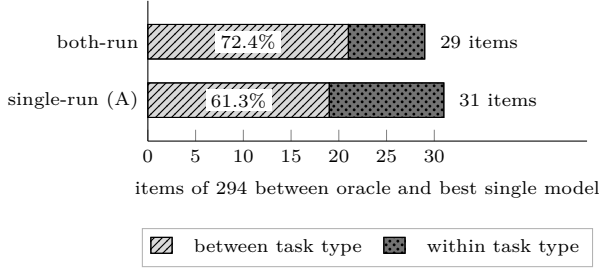

\subsection{Where the \GapStableAll items sit}\label{sec:decomp:3}

\begin{table*}[t]
\centering\footnotesize
\setlength{\tabcolsep}{5pt}
\caption{The seven scoring variants, with the main estimate marked. The distance is oracle minus best single model, given as a difference in accuracy, so 0.1054 is 10.54 percentage points. Models are named by the short names of Table 1. The four exclusion variants are the cap\_hit-excluded and parse\_failed-excluded rows. They score models on per-model denominators, which the exclusion changes, and are not \Nitems-item rankings.}
\label{tab:s3s4-1}
\begin{tabularx}{\textwidth}{@{}l>{\hsize=0.766\hsize\raggedright\arraybackslash}Xl>{\hsize=1.234\hsize\raggedright\arraybackslash}Xr@{}}
\toprule
\hd{role} & \hd{variant} & \hd{oracle} & \hd{best single model} & \hd{gap} \\
\midrule
sensitivity & single-run A $\times$ all items & \OracleNaiveA/\Nitems = 0.9524 & \texttt{claude\_fable\_5} = \texttt{claude\_opus\_5} \BestNaiveA/\Nitems = 0.8469 & 0.1054 \\
sensitivity & single-run B $\times$ all items & \OracleNaiveB/\Nitems = 0.9422 & \texttt{claude\_opus\_5} \BestNaiveB/\Nitems = 0.8571 & 0.0850 \\
main & both-run $\times$ all items & \OracleStableAll/\Nitems = 0.9320 & \texttt{claude\_opus\_5} \BestStableAll/\Nitems = 0.8333 & 0.0986 \\
sensitivity & single-run A $\times$ cap\_hit-excluded & \OracleNaiveA/\Nitems = 0.9524 & \texttt{qwen3\_7\_plus} \LangOnlyStable/282 = 0.8794 & 0.0729 \\
sensitivity & both-run $\times$ cap\_hit-excluded & \OracleStableAll/\Nitems = 0.9320 & \texttt{qwen3\_7\_plus} 238/\OracleNaiveA = 0.8500 & 0.0820 \\
sensitivity & single-run A $\times$ parse\_failed-excluded & \OracleNaiveA/\Nitems = 0.9524 & \texttt{claude\_fable\_5} \BestNaiveA/290 = 0.8586 & 0.0938 \\
sensitivity & both-run $\times$ parse\_failed-excluded & \OracleStableAll/\Nitems = 0.9320 & \texttt{gemini\_3\_7\_flash} 244/288 = 0.8472 & 0.0848 \\
\bottomrule
\end{tabularx}
\end{table*}

The distance splits in two along the task-type axis. An item in it is \textbf{between-type} if giving each task type its own best model recovers it. It is \textbf{within-type} if it survives that, because recovering it would take a different model inside its own task type. On single-run A, the 31-item distance between the oracle and the best single model (\OracleNaiveA $-$ \BestNaiveA) splits into 19 between-type items and 12 within-type items. A task-type selector therefore captures 19 of those 31 items, \ShareNaive. Of the 12 within-type items, 9 are coding. The other six task types together contribute \OtherSixNaive of \Nitems items = \OtherSixNaiveRate, below both accuracy references of \Cref{sec:noise}, so this setup does not resolve them. Those references are the \Smokerate per-item one, which is on its own scale, and the \Fliprate flip rate. The flip rate is carried on cells rather than items. The corresponding both-run split is \GapStableAll = \BetweenStable between + \WithinStable within, with 7 of the 8 within-type items in coding and 1 in instruction.

A language-only selector captures 5 of \Nitems items on single-run A (\BestNaiveA{}$\rightarrow$\LangOnlyNaive) and 3 of \Nitems under the both-run rule (\BestStableAll{}$\rightarrow$\LangOnlyStable). Inside a single language the distance is larger than that gain suggests. The gap between the oracle and that language's own best single model is 7 to 10 items of the 98 items in the language, or 7.14\% to 10.20\%. That holds on both scoring rules and in all three languages. Those are rates on 98 items, read against references carried on \Nitems items and on \Ncells cells. They clear both, so the procedure of \Cref{sec:noise:2} resolves them. The best single model also differs from one language to the next. What is small is the global gain from adding language after task type. It is 3 of \Nitems items on the single-run reading and 2 of \Nitems on the both-run one. Coding is the dominant within-type residual in ko, en, and hi; the magnitudes are not equal (single-run within coding 3/2/1; both-run 2/2/1). Non-coding per-language residuals are 0--2 items.

Choosing a per-group maximum on the same items the rule is then read on is what this paper calls the in-sample argmax. Each of Table 4's first four rules is fitted by partitioning the \Nitems items into the groups that rule's column names. Each group then goes to the model with the most correct items in it. Those groups are the whole matrix, one language, one task type, or one task$\times$language cell.

\begin{table*}[t]
\centering\footnotesize
\setlength{\tabcolsep}{5pt}
\caption{What static structure recovers on this matrix, in items out of \Nitems. The adopted lookup column is the \Cref{sec:lookup} policy, shown for reference; it is not the in-sample argmax of its row.}
\label{tab:s3s4-2}
\begin{tabularx}{\textwidth}{@{}lrrr>{\hsize=1.000\hsize\raggedleft\arraybackslash}Xlr@{}}
\toprule
\hd{rule} & \hd{best single model} & \hd{language-only} & \hd{task-type-only} & \hd{task$\times$language argmax} & \hd{adopted lookup} & \hd{oracle} \\
\midrule
single-run A & \BestNaiveA & \LangOnlyNaive & 268 & \ArgmaxNaive & n/a & \OracleNaiveA \\
both-run (A$\cap$B) & \BestStableAll & \LangOnlyStable & \TaskOnlyStable & 268 & \LookupCorrect & \OracleStableAll \\
\bottomrule
\end{tabularx}
\end{table*}

Task-type-only residual is \ResidualTaskOnlyNaive of \Nitems items on single-run A and \ResidualTaskOnlyStable of \Nitems under the both-run rule. After task$\times$language argmax the both-run residual is \ResidualArgmax of \Nitems items = \ResidualArgmaxRate. Those 6 items are also an in-sample residual, since the argmax that leaves them was fitted on the same \Nitems items, as were the \ShareNaive and \ShareStable above (\Cref{sec:limits:1}). That residual carries the comparison this paper is built around, and the reading it supports is narrow. That \ResidualArgmaxRate of items is below both accuracy references of \Cref{sec:noise}, so the procedure of \Cref{sec:noise:2} does not resolve it from what re-execution moves. The first of those references is the \Smokerate per-item one, which is on its own scale. The second is the \Nflips of \Ncells model-query cells = \Fliprate that re-executing the identical configuration moves, on a denominator of cells rather than items. The single-run residual, \ResidualArgmaxNaive of \Nitems = \ResidualArgmaxNaiveRate, sits exactly at the per-item reference and is unresolved on the same reading.

The lookup \Cref{sec:lookup} adopts is \LookupCorrect of \Nitems, not 268. It is the policy obtained after two post-result decisions defined in \Cref{sec:lookup}. The first is an override applied to the three coding cells, which moved one of them. The second is a rule that treats an accuracy margin whose sign does not hold across the two runs as a tie, and breaks it by cost (\Cref{sec:lookup:1}). It is therefore not the in-sample argmax, and it is not substituted back into this decomposition. Its own residual of 12 of \Nitems items is a \Cref{sec:lookup} figure.

The passed distance is mostly static task-type structure plus a coding residual. That is not a claim that a learned router is useful.

The gate passed and no learned router was built. It is a floor and not an instruction: it says to build only if the distance clears 5 percentage points of items (\Cref{sec:decomp:1}), so clearing it removed the bar without settling the question. What settled it was the decomposition above. Most of the distance is task type, and once task type and language are fixed the residual is 6 of \Nitems items, which \Cref{sec:noise:2} does not resolve from what re-execution moves. That reading was taken with the results in view, after the matrix was scored. \Cref{sec:limits:9} states what the reference behind it rests on.

\section{On this matrix, which model is best depends on the scoring rule}\label{sec:rule}

Which of the 14 models is the best single model on this matrix has no one answer. It depends on the scoring rule, and ten defensible rules put five different models on top. Seven of the ten are the scoring variants \Cref{sec:decomp} read along one axis, one as its main estimate and six as sensitivity checks on one distance. Read along the other axis, those seven answer a different question, which single model is best. They do not agree, on four identities. The other three rules are the main variant restricted to one language at a time, and they are the last three rows below. The fifth identity is the one that holds only the English stratum.

\begin{table*}[t]
\centering\footnotesize
\setlength{\tabcolsep}{5pt}
\caption{Who is best single model under each scoring variant, and by how much. Margins are in items where the two models share a denominator and in accuracy points where they do not.}
\label{tab:s5-scoring-rule-1}
\begin{tabularx}{\textwidth}{@{}>{\hsize=0.806\hsize\raggedright\arraybackslash}X>{\hsize=1.097\hsize\raggedright\arraybackslash}X>{\hsize=1.097\hsize\raggedright\arraybackslash}Xl@{}}
\toprule
\hd{variant} & \hd{best single} & \hd{runner-up} & \hd{margin} \\
\midrule
single-run A $\times$ all items & \texttt{claude\_fable\_5} = \texttt{claude\_opus\_5} \BestNaiveA/\Nitems & \texttt{claude\_sonnet\_5} = \texttt{qwen3\_7\_plus} \LangOnlyStable/\Nitems & tie at the top \\
single-run B $\times$ all items & \texttt{claude\_opus\_5} \BestNaiveB/\Nitems & \texttt{claude\_fable\_5} 244/\Nitems (four-way tie) & 8 items \\
both-run $\times$ all items (main) & \texttt{claude\_opus\_5} \BestStableAll/\Nitems & \texttt{gemini\_3\_7\_flash} 244/\Nitems & 1 item \\
single-run A $\times$ cap\_hit-excluded & \texttt{qwen3\_7\_plus} \LangOnlyStable/282 & \texttt{claude\_sonnet\_5} \LangOnlyStable/290 & 0.0243 \\
both-run $\times$ cap\_hit-excluded & \texttt{qwen3\_7\_plus} 238/\OracleNaiveA & \texttt{claude\_opus\_5} \BestStableAll/291 & 0.0081 \\
single-run A $\times$ parse\_failed-excluded & \texttt{claude\_fable\_5} \BestNaiveA/290 & \texttt{claude\_opus\_5} \BestNaiveA/\Nitemscapped & 0.0088 \\
both-run $\times$ parse\_failed-excluded & \texttt{gemini\_3\_7\_flash} 244/288 & \texttt{claude\_opus\_5} \BestStableAll/\Nitemscapped & 0.0110 \\
both-run, Korean only & \texttt{claude\_opus\_5} 83/98 & \texttt{gemini\_3\_7\_flash} 82/98 & 1 item \\
both-run, English only & \texttt{gpt\_5\_6\_sol} 82/98 & \texttt{claude\_opus\_5} 81/98 & 1 item \\
both-run, Hindi only & \texttt{gemini\_3\_7\_flash} 83/98 & \texttt{gemini\_3\_1\_pro} 82/98 & 1 item \\
\bottomrule
\end{tabularx}
\end{table*}

All four exclusion-variant margins are decided entirely by the denominator. Margins are items where the denominator is shared. They are accuracy points where the exclusions make it unequal. Each exclusion variant shrinks a model's denominator by that model's own excluded items. Those four margins are therefore differences between ratios with different denominators, and their columns are not \Nitems-denominator rankings. Qwen3.7 Plus and Claude Sonnet 5 both score \LangOnlyStable correct under the single-run cap-excluded rule, but drop 12 and 4 capped items respectively. Claude Fable 5 and Claude Opus 5 both score \BestNaiveA under the single-run parse-excluded rule while dropping 4 and 1. In the other two the leader holds fewer correct items than the runner-up and leads on the ratio. Those two are 244 against \BestStableAll under the both-run parse-excluded rule, and 238 against \BestStableAll under the both-run cap-excluded rule. In the second, Qwen3.7 Plus leads on \OracleNaiveA items against Claude Opus 5 on 291. Five identities appear across the ten rows: Claude Fable 5, Claude Opus 5, Qwen3.7 Plus, Gemini 3.7 Flash, GPT-5.6 Sol.

\subsection{The margins are smaller than what re-execution moves}\label{sec:rule:1}

Reading a tied top as a margin of zero, no top-two margin in Table 5 exceeds 0.0272 in accuracy. Six of the ten rows compare two models scored on the same denominator, and those six are read under the procedure \Cref{sec:noise:2} states. The largest of them is the 8-item single-run B margin, 8/\Nitems = \TopTwoRate. That is below the \Smokerate per-item reference of \Cref{sec:noise} and well below its \Fliprate model-query-cell flip rate, so the procedure leaves all six unresolved. The other four compare ratios on per-model exclusion denominators, which is not an item rate the procedure admits (\Cref{sec:noise:2}, \Cref{sec:decomp:1}). Their margins, 0.0243 at the largest, are reported as sizes rather than read against a reference. Five of the ten margins are a tie or a single item; the tied single-run A row is one item clear of third place. The two references are not the same estimand: \Smokerate is a per-item rate on \Nitems items and \Fliprate is a model-query-cell flip rate on \Ncells.

The same point shows up inside one model. Opus scores \BestNaiveA/\Nitems on run A and \BestNaiveB/\Nitems on run B, a 3-item move on identical inputs, which is larger than the 1-item margin that decides the main variant. Under the main variant the leader is \BestStableAll/\Nitems and the next six models are 244, 243, \SolCorrect/\Nitemscapped, 238, 238, and 236. So seven of fourteen sit within the \Smokerate reference of the leader, the seventh exactly at it, and nine sit within \Fliprate. The eighth and ninth, 3.74\% and 4.42\% behind, sit between the two references. That is the one band where the two disagree. There \Cref{sec:noise:2} takes the unresolved side. The count is given under each reference rather than as one number for that reason. One of those seven, GPT-5.6 Sol, is compared on \Nitemscapped items because of an infrastructure null, so that comparison too is between ratios with different denominators. \Cref{sec:setup:4} shows both denominators wherever a null occurs, and two models are affected here. GPT-5.6 Sol is \SolCorrect of \Nitemscapped = 81.57\% under the scoring \Cref{sec:setup:4} defines, which excludes the null. It is \SolCorrect of \Nitems = 81.29\% under the diagnostic field the artifacts also carry, which counts the unscored item as incorrect. GPT-5.6 Luna is \LunaCorrect of \Nitemscapped = 76.11\% and \LunaCorrect of \Nitems = 75.85\% the same way. Neither reading changes either model's rank, and the cluster counts above compare correct-item counts, so they do not move with the denominator either. The rankings in this section are quoted on \Nitemscapped and nothing in it turns on the choice.

So the reading we can support is the weak one. On this matrix the scoring rule does not favour anyone in particular. It picks a winner out of a cluster of models that are separated by less than our measurement moves on its own. ``Which model is best'' is not a question this matrix answers. It answers ``which model is best under a stated scoring rule,'' and the rule has to be stated because it is doing the deciding.

\subsection{Two things that are not scoring-rule bias}\label{sec:rule:2}

Two patterns in Table 5 could be read as a scoring rule systematically favouring particular models, and neither supports that reading.

First, the cap\_hit exclusion promotes Qwen3.7 Plus, which has the most capped items (12 on run A, 13 on B, against 10 and 9 for DeepSeek V4 Pro). The cap is our parameter, not a property of the scoring rule. Excluding capped items asks whether that pile-up is an ability ranking or a ranking of who ran into our token limit. We report both variants rather than deciding.

Second, the both-run rule promotes Gemini 3.7 Flash, which flips zero correct bits between the two runs. Claude Sonnet 5 and DeepSeek V4 Flash each lose 16 model-query cells to 1$\rightarrow$0 flips. That is what the both-run rule is defined to do, since it requires the same model to be correct twice. It is a definitional property, not a bias discovered in the data. Both-run rankings are partly determinism rankings, and we say so rather than treating them as pure ability.

Neither case licenses the stronger claim that a scoring rule is systematically unfair to particular models. We have no evidence for that and do not make it, and we offer no account of why any individual model does better under one rule than another.

\subsection{Related observations}\label{sec:rule:3}

An earlier study on a different task has the same shape \citep{absencedetectiontax}. There, enum-constrained decoding changes measured abstention ability. In both cases the reported quantity moves with a choice that is usually recorded as a methods detail.

\section{\texorpdfstring{A static task$\times$language lookup}{A static taskxlanguage lookup}}\label{sec:lookup}

The policy this pilot adopted is a static table, not a learned router, and this section is how it was built and what it scores. \Cref{sec:decomp} had located the distance to the oracle in task type, with a residual concentrated in coding. The policy that follows is therefore a table: for each of the 21 task$\times$language cells, pick one model in advance and send every query in that cell to it. Selection maximizes both-run accuracy (the same model correct in both replicate runs), and among accuracy ties minimizes the A/B mean billed cost of the 14 items in the cell. Cost is billed \texttt{usage.cost} only; an item with a missing cost is excluded from the cost coordinate while the model is retained, with no catalog imputation.

Two post-result decisions move the unconstrained in-sample argmax over those 21 cells, both taken with the outcome table in view.\footnote{Both were authorized on \mbox{2026-08-23}, the coding hold at 03:33Z and the rule of \Cref{sec:lookup:1} at 04:44Z, and both are in the project's decision records. The order is what the argument uses; the times are given so that the record can be matched to it.} That argmax is 268/\Nitems. The first decision settled the three coding cells that had been set aside for review. Cells are named task-language throughout this section, with the language written as in the tables: \texttt{ko}, \texttt{en} and \texttt{hi} for Korean, English and Hindi. So coding-hi is the coding task type in Hindi. Only coding-hi moved. Its argmax was Claude Fable 5 on an unsigned +1 both-run edge (A +1, B +0). That edge carried a \$3.354841 premium over Grok 4.3 for the cell (\$3.449535 versus \$0.094694), so Grok 4.3 was selected instead. We call that switch the coding-Hindi override. It was a decision on those three cells and not a re-sweep of the other 18. It produced an intermediate policy scoring \UniqueWinnerCorrect of \Nitems items at a mean billed cost of \UniqueWinnerCost per run. Policy coordinates are written accuracy @ cost from here on, so that one is \UniqueWinnerCorrect/\Nitems @ \UniqueWinnerCost.

\subsection{The unsigned-margin tie rule and why it is applied uniformly}\label{sec:lookup:1}

The coding-Hindi decision above generalizes into the rule that produced the adopted table. An accuracy edge whose sign does not hold across A and B is consistent with the model-query-cell flip rate of \Cref{sec:noise} rather than with a reproducible difference. Paying a premium for it buys noise. Stated as a rule, it has two conditions. First, the top-scoring model or models in a cell lead the cheapest candidate one item behind them, by a margin whose sign does not hold across the two runs. Second, holding that lead costs more than taking the one-behind candidate. Where both hold, the margin is treated as a tie and the tie is broken by minimum billed cost. We call this the unsigned-margin tie rule. It is stated over the top of the cell and not over a unique winner. Five of the six cells whose selection it decides have several models tied at the top. Restricting it to unique winners is one of the readings priced and rejected below.

The rule's domain was ambiguous, and Table 6 prices the three readings as they stood before one was chosen.

\begin{table*}[t]
\centering\footnotesize
\setlength{\tabcolsep}{5pt}
\caption{The three readings of the tie rule's domain, priced on the same matrix before one was adopted. Accuracy is both-run correct items out of \Nitems and the cost is the policy's mean billed cost per run over the \Nitems-query workload. The three readings are lettered A, B and C in the artifacts; this section names them, since A and B already name the two runs.}
\label{tab:s6-lookup-1}
\begin{tabularx}{\textwidth}{@{}>{\hsize=0.509\hsize\raggedright\arraybackslash}X>{\hsize=1.491\hsize\raggedright\arraybackslash}Xrr@{}}
\toprule
\hd{reading} & \hd{domain} & \hd{both-run accuracy} & \hd{mean billed} \\
\midrule
unique-winner (rejected) & only cells with a unique both-run winner at a one-item margin & \UniqueWinnerCorrect/\Nitems & \UniqueWinnerCost \\
cheapest-one-behind (adopted) & every cell whose unsigned margin over the cheapest one-behind carries a price premium & \LookupCorrect/\Nitems & \LookupCost \\
wide-pool (rejected) & tie pool widened to all one-behind models with unsigned margins & \WidePoolCorrect/\Nitems & \WidePoolCost \\
\bottomrule
\end{tabularx}
\end{table*}

The unique-winner reading was rejected because it applies the rule to two named cells while leaving other cells with the same property untouched, which is selection after seeing the result. Its accuracy and cost coincide with the intermediate the coding decision produced, because neither named cell moves under it. The cell coding-hi was already switched by that decision, and toolcall-hi is already the minimum-billed candidate. The wide-pool reading was rejected because widening the tie pool beyond the cheapest one-behind has no basis in the rule as stated. It would be a new post-result rule, and it gives up one more item, in instruction-ko. There it replaces DeepSeek V4 Pro with Grok 4.3 and lowers the policy's mean billed cost by \$0.006157. The cheapest one-behind in that cell carries a signed margin, so it is the widening of the pool, and not the rule as adopted, that moves the cell. The cheapest-one-behind reading was adopted.

The adopted reading is a single pass over the pre-tie-rule argmax table and does not recurse after replacement. A second-pass audit finds zero further applicable cells, so the single-pass restriction does not hide any candidate.

The rule filters nothing when the unsigned winner is already the minimum-billed candidate. The cell toolcall-hi is that case. Gemini 3.7 Flash holds an unsigned +1 (A +1, B +0) over DeepSeek V4 Pro but is cheaper (\$0.008287 vs \$0.009676), so it stays selected. That is intended. The rule blocks premiums; it does not penalize cheap candidates. Every cell of the adopted table carries a one-line record of what its selection rests on, written when the table was built and kept with it. We call that the cell's evidence grade. The toolcall-hi grade reads ``tie-break selection; no signed accuracy premium,'' which no other cell carries. It is not the only thin selection. Six further cells read ``rule-b uniform premium tie-break; no signed accuracy premium,'' where ``rule-b'' is the label the artifacts give the unsigned-margin tie rule. So seven of the 21 cells rest on no signed accuracy margin. Counted the other way, only 5 of the 21 selected models are the unique highest-scoring model in their cell under the both-run rule. Another 10 tie at the top and are picked on cost. The remaining 6 sit one item below their cell's best, because the unsigned-margin tie rule or the coding-Hindi override took the cheaper one-behind candidate.

\subsection{What uniformity costs}\label{sec:lookup:2}

Applying that rule to every cell it covers, rather than only where it was noticed, costs accuracy. Beyond coding-hi it changes exactly five of the 21 cells, each losing one item, and Table 7 names them.

\begin{table*}[t]
\centering\footnotesize
\setlength{\tabcolsep}{5pt}
\caption{The five task$\times$language cells that uniform application changes, each losing one item of \Nitems.}
\label{tab:s6-lookup-2}
\begin{tabularx}{\textwidth}{@{}llll>{\hsize=1.000\hsize\raggedright\arraybackslash}X@{}}
\toprule
\hd{cell} & \hd{before} & \hd{after} & \hd{both-run correct} & \hd{mean billed before $\rightarrow$ after} \\
\midrule
math-ko & gpt\_5\_6\_luna & deepseek\_v4\_flash & 14$\rightarrow$13 & \$0.002497 $\rightarrow$ \$0.001020 \\
extraction-hi & gpt\_5\_6\_terra & gpt\_5\_6\_luna & 14$\rightarrow$13 & \$0.008242 $\rightarrow$ \$0.001429 \\
toolcall-ko & deepseek\_v4\_pro & deepseek\_v4\_flash & 13$\rightarrow$12 & \$0.007751 $\rightarrow$ \$0.000829 \\
abstention-ko & gpt\_5\_6\_luna & deepseek\_v4\_flash & 14$\rightarrow$13 & \$0.007293 $\rightarrow$ \$0.001115 \\
abstention-en & grok\_4\_3 & deepseek\_v4\_flash & 14$\rightarrow$13 & \$0.014978 $\rightarrow$ \$0.001105 \\
\bottomrule
\end{tabularx}
\end{table*}

The unsigned-premium pattern that motivated the rule in coding-hi is not a coding-specific property. That is the argument for applying the rule uniformly rather than at the cells where it was noticed. The five cells are spread across four task types (math 1, extraction 1, toolcall 1, abstention 2) and across languages 3 ko / 1 en / 1 hi. None is coding.

The procedure of \Cref{sec:noise:2} leaves the loss unresolved. It is reported at its size and in its direction, and this setup cannot separate it from what re-execution moves. The 5-item loss is \UniqueWinnerCorrect{}$\rightarrow$\LookupCorrect, or \UniformLoss of \Nitems items = \UniformLossRate. It is smaller than both accuracy references of \Cref{sec:noise}, though the three quantities have different denominators. Both \UniformLossRate and the \Smokerate per-item reference are rates on \Nitems items, while \Fliprate is a model-query-cell flip rate on \Ncells. Every one of the five edges is unsigned across A and B. That is the price of a rule with no cell-specific exceptions, and it is reported as a loss rather than absorbed into the headline.

\subsection{The adopted lookup}\label{sec:lookup:3}

The table that these two decisions produce has a both-run accuracy of \textbf{\LookupCorrect/\Nitems = \LookupAcc} at a mean billed cost of \textbf{\LookupCost} (A \LookupCostA, B \LookupCostB) for the \Nitems-query workload. Residual to the both-run oracle \OracleStableAll is 12 items, of which 6 are coding. It selects Gemini 3.7 Flash in 6 cells, DeepSeek V4 Flash in 6, GPT-5.6 Luna in 4 and Grok 4.3 in 2. Claude Opus 5, DeepSeek V4 Pro and Claude Fable 5 take 1 each. Seven of the fourteen candidate models appear.

Cost is concentrated: coding contributes \CodingCost of \LookupCost = \CodingCostShare, and coding-ko on Claude Fable 5 alone is \CodingKoCost = \CodingKoShare of the total. That single cell also drives the lookup's own A/B billed spread of \LookupSpread through its \CodingKoSpread movement between runs, so the mean is a mean over a noisy axis. The cell coding-ko is the one expensive cell whose edge is signed (both-run +3, A +2, B +4).

\begin{figure}[t]
\centering
%
\begin{tikzpicture}
\begin{axis}[
  width=\linewidth, height=6.0cm,
  xmode=log, log basis x=10,
  xlabel={mean billed cost per run, USD (log scale)},
  ylabel={items correct of \Nitems},
  xlabel near ticks, ylabel near ticks,
  xmin=0.075, xmax=26, ymin=204, ymax=278,
  ytick={210,220,230,240,250,260,270},
  grid=both, grid style={draw=black!10},
  tick label style={font=\scriptsize},
  label style={font=\scriptsize},
  legend style={font=\scriptsize, at={(0.5,-0.32)}, anchor=north,
                legend columns=2, draw=black!30, column sep=6pt},
  clip mode=individual,
]

\draw[->, >=stealth, line width=0.7pt, draw=black!70, dashed]
  (axis cs:6.151118,246.5) -- (axis cs:4.166137,260.5);

\addplot[only marks, mark=o, mark size=1.9pt, draw=black!60]
  coordinates {(14.461315,238) (4.406287,232) (0.107520,210) (0.831765,227) (7.024294,243) (0.469347,244) (0.171896,223) (1.288316,239) (1.523484,234) (0.697917,218) (3.112309,220) (7.717874,236) (1.073490,238)};
\addlegendentry{single model}

\addplot[only marks, mark=square*, mark size=3.1pt, draw=black,
         fill=black!45]
  coordinates {(7.688898,245)};
\addlegendentry{best single model}

\addplot[only marks, mark=star, mark size=4.6pt, line width=0.9pt,
         draw=black]
  coordinates {(3.332910,262)};
\addlegendentry{adopted lookup}

\addplot[only marks, mark=triangle*, mark size=3.1pt, draw=black,
         fill=black!10]
  coordinates {(0.542932,264)};
\addlegendentry{all-coding-Grok}

\node[font=\scriptsize, align=left, anchor=south] at (axis cs:0.542932,266.4) {all-coding-Grok};
\node[font=\scriptsize, align=left, anchor=south] at (axis cs:3.332910,264.4) {lookup};
\node[font=\scriptsize, align=left, anchor=west] at (axis cs:9.995567,246.0) {Claude\\Opus 5};
\end{axis}
\end{tikzpicture}
\caption{Billed cost against accuracy for all \Nmodels\ single models and
for the two static policies this section names, under the both-run rule.
The lookup sits up and to the left of the best single model, which is the
dominance the dashed arrow marks. The all-coding-Grok point is the frozen
pre-tie-rule version, so it and the lookup are not two points on one
frontier and none is drawn (\Cref{sec:lookup:5}). Series are distinguished
by marker shape as well as by shade.}
\label{fig:pareto}
\end{figure}
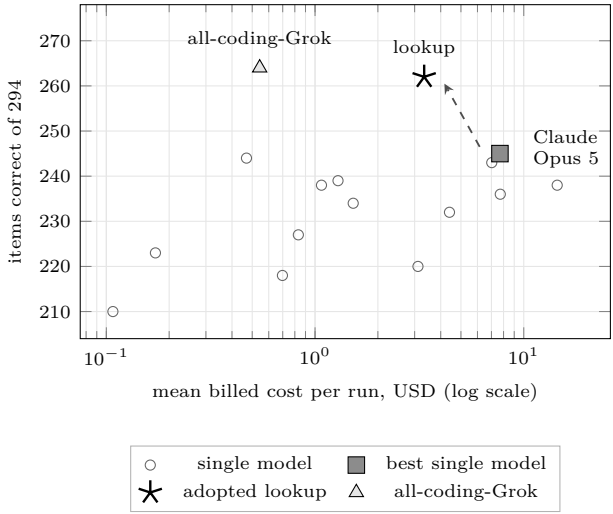

\subsection{Against the best single model}\label{sec:lookup:4}

The best single model by both-run accuracy on the same frozen \Nitems items is Claude Opus 5 at \BestStableAll/\Nitems and \OpusCost mean billed. The lookup is \textbf{+\LookupVsOpusItems items and $-$\LookupVsOpusCost} against it: ahead on accuracy and on billed cost, on an identical workload, with no per-item oracle involved. That is the comparison that matters for the adoption decision. Under single-run scoring the best single model is a different figure (\Cref{sec:decomp}), which is why the comparison names its scoring rule. The procedure of \Cref{sec:noise:2} resolves both differences, the accuracy one narrowly. The accuracy difference, 17 of \Nitems items, is 5.78\%, well above the \Smokerate per-item reference, which is on its own scale. It is also above the \Fliprate flip rate, which is carried on cells rather than items and is the reference that decides here. \Cref{sec:limits:9} gives the flip-rate estimate that would take this reading back. The cost difference is 56.7\% of what Opus bills, against cost references of \Billedmove and 14.79\%. All three readings of the tie rule, \UniqueWinnerCorrect/\Nitems, \LookupCorrect/\Nitems and \WidePoolCorrect/\Nitems, are ahead of Opus on both axes, so the conclusion does not depend on which one was adopted.

Opus's billed cost is computed over \Nitemscapped items per run, not \Nitems: its English parallel-tool-call item is \texttt{parse\_failed} in both runs (\Cref{sec:setup:4}) and has no billed cost. The item stays in the accuracy denominator at correct=0; only its cost is missing, and no catalog value is substituted. The lookup selects Gemini 3.7 Flash for toolcall-en and contains zero cost-null requests, so its own cost denominator is \Nitems.

The direction survives the run-to-run movement on both sides, not only on the mean: the lookup's costlier run bills \LookupCostA against Opus's cheaper run at \OpusCostB.

\subsection{The cheap symmetric alternative}\label{sec:lookup:5}

Being ahead of Opus is not the same as being the best static policy available. Sending all coding to Grok 4.3 and keeping the lookup's non-coding cells is far cheaper. Two versions of that policy exist, and the procedure of \Cref{sec:noise:2} is applied to both. That procedure reports a difference at or below either of the references that apply to it as unresolved by this setup, ours included.

Applying the unsigned-margin tie rule symmetrically to the 18 non-coding cells the two policies share builds it under the same version of the tie rule as the lookup. That version gives \textbf{\SymmetricCorrect/\Nitems @ \SymmetricCost}. Against it the lookup is 3 of \Nitems items more accurate and \LookupVsSymCost more expensive per run. Three of \Nitems items is \LookupVsSymRate, below the \Smokerate per-item reference of \Cref{sec:noise} and below its \Fliprate model-query-cell flip rate on its own scale. So this setup does not resolve the accuracy difference. The cost difference is \LookupVsSymCostShare of what the lookup itself bills. That is far outside the \Billeddelta cost reference of \Cref{sec:noise}, and outside the \LookupSpread run-to-run band its own cost carries (\Cref{sec:limits:7}). \Cref{sec:noise:2} resolves it. What the matrix resolves here is one difference and not two. The cheaper policy is ahead on the axis this setup can read, and adopting the lookup pays \LookupVsSymCost per run for an accuracy edge it cannot.

The second version freezes the pre-tie-rule non-coding selections and is \FrozenCorrect/\Nitems @ \FrozenCost. It is cheaper than the lookup by \LookupVsFrozenCostShare of what that policy bills, again far outside the cost reference and its own run-to-run band. It is also 2 of \Nitems items more accurate, a difference the setup does not resolve either. \Cref{sec:lookup} does not read the two as points on one accuracy/cost frontier. The two were built under different versions of the tie rule, and that is the reason, not where either sits.

The two coordinates stood differently when the coding cells were decided. The frozen one was already on the table and was set aside there by name. The decision record lists it at \FrozenCorrect/\Nitems @ \FrozenCost as the cheapest of the 27 top-three combinations. It records that those 27 span 260 to 268 of \Nitems items rather than forming a plateau, and it rules out choosing on cost alone. The symmetric one did not exist yet, having been built after the unsigned-margin tie rule was fixed. What that decision fixed is the principle that the minimum-cost combination is not adopted as the policy, which applies to the first directly and to the second by extension. Either way, preferring the lookup is a human decision on the accuracy/cost trade and not one the data forces.

The lookup was selected on the same \Nitems items it is evaluated on, and the unsigned-margin tie rule itself was fixed after the matrix was visible. Both \LookupCorrect/\Nitems and the \LookupVsOpusItems-item difference are therefore in-sample. That limitation is stated here and developed in \Cref{sec:limits}.

\section{Limitations}\label{sec:limits}

The entries below run from this paper's two results outward to the matrix both are computed on, and end at the execution floor of \Cref{sec:noise}. Those two results are the gap decomposition of \Cref{sec:decomp} and the static lookup of \Cref{sec:lookup}. That order is not a severity ranking, and no entry is labelled the main limitation. 7.1 binds the policy result and 7.2 the decomposition. 7.3 through 7.8 bind what the matrix is a matrix of: its cells, its coding items, its scoring, its model pool, its cost axis and its languages. 7.9 binds every place where the paper reads a difference against a reference.

\subsection{The policy was selected on the items it is evaluated on}\label{sec:limits:1}

The lookup of \Cref{sec:lookup} was fitted on the same frozen \Nitems items it is then scored on. It is a per-cell selection over all 14 candidates, followed by two post-result decisions. That is why it is \LookupCorrect rather than the in-sample argmax \ArgmaxStable. Pairing against Claude Opus 5 does not remove the selection bias introduced by fitting the table on those items; it fixes only the evaluated workload. \LookupCorrect/\Nitems = \LookupAcc is therefore an in-sample upper-bound estimate for the selected policy. The \LookupVsOpusItems-item difference against Opus is an in-sample comparison against a single-model baseline. It is not an upper bound, and not an out-of-sample performance claim. We apply no shrinkage, bias correction, or other adjustment to \LookupCorrect/\Nitems, to the 12-item residual, or to \LookupCost. For the residual we claim no direction of bias at all. \OracleStableAll is itself a hindsight quantity, a per-item union over 14 models taken with the answers in view. Both terms of \OracleStableAll $-$ \LookupCorrect are therefore optimistic, and this data does not say which is more so. The cost figure is taken up in the third paragraph of this entry. Why no useful holdout was available is 7.3.

There is a second layer above item reuse: the selection rule was also fixed after the matrix was visible. The unsigned-margin tie rule was adopted after the three coding cells set aside for review had already been decided (\Cref{sec:lookup}). Its domain was settled by pricing three readings against the same data and choosing one. Those readings are \UniqueWinnerCorrect/\Nitems at \UniqueWinnerCost, \LookupCorrect/\Nitems at \LookupCost and \WidePoolCorrect/\Nitems at \WidePoolCost (\Cref{sec:lookup:1}). A rule chosen with the outcome table in view is not a rule fixed in advance, and we do not present it as one. The argument for applying it uniformly is in \Cref{sec:lookup} and is not repeated here as mitigation.

The cost coordinate carries the property too. \LookupCost is the billed mean of a policy whose accuracy ties were broken toward the cheaper model on these same 14-item cells. It is an in-sample figure exactly as \LookupCorrect/\Nitems is: the cost of the table these items produced, not an estimate of what a table fitted on other items would cost. Here too we claim no direction. Resolving a tie toward the cheaper model can only lower the total within that tie. But \CodingKoShare of the total sits in a single cell selected on a signed accuracy edge rather than on cost. A re-fit that lost that edge would move the total by more than every tie-break in the table combined, in whichever direction the replacement's price runs.

The selections are also thinner than the policy total suggests. Only 5 of the 21 selections pick the unique highest-scoring model in their cell under the both-run rule. One of those five, the toolcall-hi selection, rests on an edge whose sign does not survive the second run. Ten more tie at the top of their cell and are picked on billed cost. The remaining six sit one item below their cell's best, because the unsigned-margin tie rule or the coding-Hindi override took the cheaper one-behind candidate. They are math-ko, extraction-hi, toolcall-ko, abstention-ko, abstention-en and coding-hi. Seven of the 21 cells carry an evidence grade that records the absence of a signed accuracy premium.

Being ahead of Opus is also not the same as being the best static policy on this matrix. Under the same rule version, sending all coding to Grok 4.3 gives \SymmetricCorrect/\Nitems at \SymmetricCost. That is a coordinate constructed after the fact, not a policy \Cref{sec:lookup} puts forward. Against it the lookup is 3 of \Nitems items more accurate and \LookupVsSymCost more expensive per run. Three of \Nitems items is \LookupVsSymRate, below the \Smokerate per-item reference of \Cref{sec:noise} and below its \Fliprate model-query-cell flip rate on its own scale. So this setup does not resolve the accuracy difference. The cost difference, \LookupVsSymCostShare of what the lookup bills, is far outside the cost reference of \Cref{sec:noise} and outside this lookup's own run-to-run band (\Cref{sec:limits:7}). Only one of the two differences is resolved, and it favours the cheaper policy, which is the reading \Cref{sec:lookup:5} gives. Adopting the lookup pays \LookupVsSymCost per run for an accuracy edge the matrix cannot read, and that is a human decision rather than one the data forces. The pre-tie-rule frozen version of that same policy is \FrozenCorrect/\Nitems at \FrozenCost. It is cheaper by a margin far outside the same cost references, and 2 of \Nitems items more accurate, which this setup does not resolve. \Cref{sec:lookup} excludes it from frontier comparison on rule-version grounds rather than on its coordinates.

The same in-sample property attaches outside \Cref{sec:lookup}. \Cref{sec:decomp}'s argmax selectors and the per-item oracle that defines its gap are both formed on these items. \Cref{sec:rule}'s best single model is a maximum over the fourteen models on these same items, taken on \Nitems items or on the exclusion and per-language denominators \Cref{sec:rule} states. \Cref{sec:intro} and \Cref{sec:decomp:3} both state it for the \Cref{sec:decomp} shares. \Cref{sec:rule} does not state it for its leader identities, and the abstract's closing sentence is what covers them. There it attaches to who those leaders are rather than to a claim about out-of-sample performance, since \Cref{sec:rule} claims only that the identity moves with the scoring rule. Nothing is selected in \Cref{sec:noise}.

This weakens three readings. The first is \LookupCorrect/\Nitems or the \LookupVsOpusItems-item difference as expected out-of-sample performance. The second is any claim that the lookup is the policy this matrix selects, rather than one of several defensible static policies. The third is any reading of \Cref{sec:decomp}'s \GapStableAll-item distance as the gain a router would recover on fresh items. Its oracle term is a maximum taken per item, and its best-single-model term a maximum taken over the fourteen models. It does not weaken the paired comparison as a statement about these \Nitems items, where the lookup and the Opus baseline see an identical workload on the accuracy axis. On the cost axis the denominators differ, which is 7.7.

\subsection{The task-type axis is not separable from source dataset here}\label{sec:limits:2}

Each of the seven task types draws all of its items from one source (\Cref{sec:setup:2}). On this matrix, conditioning on task type and conditioning on source are the same operation. Prior work already reports that much of the available gain may be coarse-grained domain structure, on a coordinate a router can observe (\Cref{sec:intro}). The shares of \Cref{sec:decomp} are \BetweenNaive of the \GapNaive items under one scoring rule and \BetweenStable of the \GapStableAll under the other. They are therefore consistent with a re-measurement of that coordinate on a new pool of models. They are equally consistent with task kind being the operative property. This matrix cannot tell the two readings apart. Separating them needs at least two sources inside one task type, so that a selector conditioned on the task can be compared against one conditioned on the corpus. The design has none. The seven types draw seven disjoint item sets from six sources, one of them built in-house. Only tool selection and abstention share a source, and they take disjoint categories of one release of it. That pair is one corpus split into two task types, rather than one task type drawn from two corpora.

Two facts narrow the claim without repairing it. The seven types were fixed as task kinds first, and the sources were then chosen to instantiate them, so the labels were not read off the corpora. Those kinds are multiple-choice knowledge, grade-school arithmetic, instruction following, structured extraction, tool selection, abstention, and competitive programming. And \Cref{sec:decomp}'s concentration of the residual in one type is a statement about coding items under either reading. Neither turns a confounded axis into a separated one.

The language axis does not carry this problem. One item identifier denotes the same item in all three languages (\Cref{sec:setup:3}), so language varies with the source held fixed. The per-language results of \Cref{sec:decomp} and \Cref{sec:rule} are not open to the same reading. That is the one axis of this decomposition free of this confound. It carries a different limitation of its own, which is 7.8. Two of the three conditions are one pipeline's renderings of English-authored items. The axis is free of the source confound without being a clean measurement of query language as users write it.

This weakens the first contribution as an identification claim. \Cref{sec:decomp}'s share should be read as what a selector conditioned on this matrix's seven task-and-corpus pairs captures. It is not evidence that task kind rather than corpus identity is the operative coordinate. It does not weaken the share as a measurement, since a table keyed on those seven labels does capture \BetweenNaive of \GapNaive and \BetweenStable of \GapStableAll items. It does not reach \Cref{sec:lookup}'s in-sample evaluation either, whose lookup is keyed on exactly those labels and is scored on exactly those items. It does bear on applying that table to traffic drawn from other corpora.

\subsection{\texorpdfstring{Fourteen items per task$\times$language cell}{Fourteen items per taskxlanguage cell}}\label{sec:limits:3}

There is no holdout, and at this cell size no useful one was available. Each of the 21 task$\times$language cells holds 14 items, so one item is 7.14\% of a cell. Every cell-level difference the matrix can express is a multiple of that. The exception is the one cell where an infrastructure null leaves 13 for two models, taken up below. The same 14 items are used to select and to evaluate. A selection/evaluation split leaves 7 items on each side, so neither the fitted choice nor the estimate of it would be stable.

The paper does not keep to its own rule everywhere, and \Cref{sec:setup} now points here for the exceptions. The configuration says as much at pin time. Readings are to be taken from the whole-matrix total, and 14 per cell is called insufficient even for per-task-type interpretation. \Cref{sec:setup} carries that forward as a rule for the paper: task$\times$language numbers are meant to be read as components of an aggregate. \Cref{sec:intro} states it for its own claims. \Cref{sec:decomp} locates the residual in one task type and reports a 9-item coding count, and per-language counts of 0 to 3 items in single task$\times$language slices. \Cref{sec:rule} reports per-language leaders on 98 items. \Cref{sec:lookup} fits and fixes 21 per-cell selections, one of which carries \CodingKoShare of the cost axis. Those readings are assembled from 14-item cells, and this limitation attaches to them.

Denominators are not uniform either. Three keys remained infrastructure failures after the retry budget was exhausted. They are GPT-5.6 Sol and GPT-5.6 Luna in run A, and GPT-5.6 Luna again in run B, all on the same Korean coding item. Two model-query cells are affected. An infrastructure null is neither 0 nor 1, so those two models are scored on \Nitemscapped comparable items, \SolCorrect/\Nitemscapped and \LunaCorrect/\Nitemscapped. \Cref{sec:rule} gives both readings for both models, the second charging the unscored item against them at \SolCorrect/\Nitems and \LunaCorrect/\Nitems, and records that neither model changes rank between the two. The flip rate carries the same split: \Nflips/\Ncells over all model-query cells, \Nflips/\Ncomparable over the comparable ones, \Fliprate either way. \Cref{sec:rule} marks the comparisons that are ratios between unequal denominators.

This weakens several readings. It weakens any statement about which model is better at a specific task$\times$language pair. It weakens the stability of the 21 selections, since one item changing in a cell can move the choice. It weakens the per-language leaders of \Cref{sec:rule}, which 7.8 takes up. It weakens any ratio that puts GPT-5.6 Sol or GPT-5.6 Luna against a model scored on \Nitems, which \Cref{sec:rule} marks in place. And it weakens the per-task-type readings of \Cref{sec:decomp}, which the configuration itself calls under-powered at this cell size. It does not weaken the whole-matrix aggregates that are not assembled from per-cell choices. Those are the flip rate, which is \Fliprate on either of its denominators, and the gap between the oracle and the best single model over \Nitems items. \LookupCorrect/\Nitems is not in that group. It is the sum of 21 selections made at this cell size.

\subsection{No contamination control on the coding task type}\label{sec:limits:4}

No exposure test was run on the coding items. They are LiveCodeBench \texttt{release\_v6} restricted to \mbox{2025-01-01} through \mbox{2025-04-06}, the end of that release, with \texttt{contamination\_control: none}. The window closes \mbox{2025-04-06}, sixteen months before the \mbox{2026-08-21} catalog this model pool is drawn from. That is a catalog date, not a training cutoff. The configuration records that these items are not post-cutoff data for most of the pool. The one related design step is difficulty stratification, 3 easy / 5 medium / 6 hard drawn from a 45/55/82 item pool, which bears on interpretation and not on exposure. Contamination can move the coding numbers in either direction. Items memorized by every model align the answers and shrink the task type's spread. Items memorized by part of the pool inflate the differences between models. This design cannot tell the two apart.

Coding is where that matters most, because coding carries both the residual and the cost. After a task-type selector, 9 of the 12 remaining within-type items are coding under the single-run rule and 7 of the 8 under the both-run rule. A full task$\times$language argmax leaves 6 items, 5 of them coding (ko 2, en 2, hi 1) and the sixth instruction-ko. After the two post-result decisions that produce the adopted policy, 6 of its 12 residual items are coding. On the cost axis coding is \CodingCost of the lookup's \LookupCost billed mean, \CodingCostShare, with the single coding-ko cell at \CodingKoShare.

This weakens the finding that the residual left after task type and language is concentrated in coding. Memorization by part of the pool produces model differences on this task type that are not ability differences, which is the same shape as the result we report. It weakens the three coding selections in the lookup, and with them the exact size of the \LookupVsOpusItems-item difference. It weakens the whole cost story too, \CodingKoShare of which is the single coding-ko cell. It reaches \Cref{sec:rule} only in part. The mechanism behind the split in identities is that the scoring variants differ in denominators and in how they treat run-to-run disagreement, and neither is contamination-driven. But which identities appear at the top, and whether the variants disagree at all, can move with contamination that reaches part of the pool. Coding is 42 of the \Nitems items those rankings are computed on, and 14 of the 98 in each per-language ranking. Five of \Cref{sec:rule}'s ten margins are a tie or one item. It does not weaken the flip rates of \Cref{sec:noise}, which compare a model against itself on identical items.

\subsection{``Correct'' here means a rule-based binary check}\label{sec:limits:5}

All seven scorers are rule-based and binary: multiple choice, exact numeric match, IFEval constraint checking, exact JSON key-value match, tool-call match on the parsed call's abstract syntax tree, abstention as the absence of any parsed call, and executed unit tests. Models do produce free-form output. The adopted IFEval types include \texttt{combination:two\_responses} and \texttt{length\_constraints:number\_paragraphs}, and the coding task type runs generated programs. But nothing is graded on content quality: there is no model judge and no human rating anywhere in the matrix. An item is scored 1 or 0, so nothing here separates a minimally passing answer from a good one. The oracle is the union of models that produced a checkably correct output rather than the union of models that produced the best one.

This weakens any transfer of the \ShareNaive and \ShareStable task-type shares, of the oracle gap, or of the lookup to work whose quality is not a rule-checkable binary. That kind of work includes open-ended writing, summarization, long-context work and multi-turn dialogue, none of which this matrix scores. We have no measurement of the traffic mix here. Work of those kinds may be part of what a production router would be asked to route. To that extent the policy of \Cref{sec:lookup} is unevaluated on it, and for that traffic the routing question is not answered here in either direction. It does not weaken the results as statements about rule-checkable correctness on the items the matrix does contain, where the scoring rule is stated and applied identically to every model. It does weaken any reading of them as statements about answer quality.

\subsection{One catalog snapshot, one gateway}\label{sec:limits:6}

Every headline quantity is relative to the pool. That pool is 14 models from a \mbox{2026-08-21} catalog snapshot, identifiers verified by live call on \mbox{2026-08-22}, all reached through a single gateway. The oracle is a union over these 14 models, so a larger pool cannot lower it. But a larger pool can also raise the best single model, so the gap between them may widen or narrow. The \ShareNaive and \ShareStable shares are ratios of between-type items to total distance items, both of which move with the pool in ways this data cannot predict.

The policy is pool-bound in a second way. Each of the 21 cells is fitted against all 14 candidates. Adding or removing one model means a fresh pass over all 21 cells, rather than only the cells the new model would win. The unsigned-margin tie rule breaks ties toward the cheapest one-behind candidate, and that candidate can change anywhere. Running through one gateway also bundles serving-side behaviour into the identity of a model. The probe recorded in \Cref{sec:setup} confirms that the routed model equalled the requested model, which is not the same as confirming what weights served it.

This weakens the transportability of every number here to another pool, a later catalog date, or a different provider path. It also weakens the durability of the 21 selections. It does not weaken the comparisons made inside the matrix as comparisons between served endpoints, since all 14 were reached the same way in the same two runs. It does weaken any reading of them as comparisons between sets of weights.

\subsection{The cost axis is billed, and billed moves}\label{sec:limits:7}

The cost axis is what this account was charged on two runs, and no figure on it is a price statement. Cost is the gateway's billed \texttt{usage.cost}, pinned before any of the 14-model scores were seen. The three available figures have been compared once, on an earlier one-model trial run rather than on this matrix. They disagree pairwise: billed \$0.044562, the configured catalog reconstruction \$0.056087, the gateway's list reconstruction \$0.032505. That comparison was not repeated on the 14-model matrix. Apart from those two reconstructions, we report billed amounts only and do not mix them with a price table.

Billed cost also moves between identical runs: \Billeddelta at the matrix level and \LookupSpread for the lookup, the latter driven by coding-ko's \CodingKoSpread movement. The matrix-level figure understates the movement for a policy whose cost concentrates in one cell, which is the shape of this lookup.

Eight of the fourteen models' recorded prices are conditional, expiring, freshly cut, or not the prices this gateway lists. Prices here are USD per million tokens and are written input/output, as in Table 1. Two are introductory and expire. Claude Sonnet 5's 2.00/10.00 becomes 3.00/15.00 on \mbox{2026-09-01}. Gemini 3.7 Flash's 0.75/3.75 becomes 1.50/7.50 on \mbox{2027-01-01}, with the gateway listing 0.375/1.875 for it in the meantime. Two meter differently on prompts above 200K tokens, Gemini 3.1 Pro and Grok 4.6. Two are configured at the vendor's direct-API prices rather than the gateway's. For DeepSeek V4 Pro it lists 1.188/3.564 against the configured 0.435/0.87, which is 2.7$\times$ on input and 4.1$\times$ on output. For DeepSeek V4 Flash the relation runs the other way: it lists 0.0679/0.168 against the configured 0.14/0.28, so the configured price is 2.1$\times$ and 1.7$\times$ the listed one. Two were cut shortly before the snapshot: GPT-5.6 Terra and GPT-5.6 Luna had input prices cut 20\% and 80\% on \mbox{2026-07-30}, three weeks earlier. The only place any of these values is used is the reconstruction above; no accuracy or cost-axis figure is computed from them.

They bear unevenly on the policy. Four of those eight models are ones the lookup selects, and they hold 17 of the 21 cells. Those four are Gemini 3.7 Flash, DeepSeek V4 Pro, DeepSeek V4 Flash and GPT-5.6 Luna. Sixteen of those 17 were settled by a cost comparison, so a table re-fitted at other prices could move up to 16 of the 21 selections. The comparison was either an accuracy tie broken on billed cost, or the unsigned-margin tie rule treating an unsigned edge as a tie and breaking it the same way. Only toolcall-en among them rests on an accuracy edge that no price moves. That bound is a count of conditioned prices, not a measurement. The rule minimizes billed cost, and how a change in a recorded price maps onto a billed amount is exactly what the first paragraph's three-way disagreement leaves open. Those same 17 cells are \$0.166446 of the \LookupCost mean, 5.0\%. The \CodingKoShare sits on Claude Fable 5 in one coding cell, whose recorded price carries no condition. The same is true of the prices behind the three cells the table fills with Grok 4.3, twice, and Claude Opus 5, once. The exposure is therefore uneven across the two axes of the same policy. Sixteen selections rest on a cost comparison that involves a conditioned price, while 95.0\% of the measured total sits on prices that carry none.

One model-query cell has no billed cost at all. Claude Opus 5's English parallel-tool-call item is \texttt{parse\_failed} in both runs (\Cref{sec:setup:4}) and returned no billed amount. Each Opus run total is therefore a sum over \Nitemscapped billed items, and \OpusCost is the mean of those two totals, while its accuracy denominator stays \Nitems. Nothing is imputed, so \Cref{sec:lookup} puts a \Nitems-item cost against a \Nitemscapped-item one.

This weakens the $-$\LookupVsOpusCost difference as a durable price statement, and any cost ordering read forward past the introductory windows. It weakens the precision of \LookupCost, whose run-to-run band is the \LookupSpread spread rather than the matrix-level \Billeddelta. It does not weaken the direction of the \Cref{sec:lookup} cost comparison, which holds run by run. The lookup's costlier run bills \LookupCostA against Opus's cheaper one at \OpusCostB, and the unbilled item can only add to Opus's side.

\subsection{Three languages, and two of them are translations}\label{sec:limits:8}

What varies across the language axis is translation-induced variation on English-authored items, not a native query distribution. The language conditions are Korean, English and Hindi over one frozen 98-item ledger. Knowledge uses MMLU-ProX's existing parallel release. The other six task types were translated from English by one pipeline at temperature 0 with seed 42. Numbers, code, identifiers, schemas and instruction arguments were held out of translation. Register, length and topical priors are inherited from the English source rather than drawn from native traffic, and the pipeline's own failure modes sit inside the language condition itself.

Because one ledger serves all three languages, an item dropped on the strength of its Hindi rendering is dropped from English and Korean too. Three of the 14 IFEval items, 21.43\%, failed first-pass validation in Hindi. Two were replaced and one was retranslated once and passed, so no item in the final ledger fails validation. The instruction task type in every language is therefore conditioned on Hindi-translation survival. The Hindi instruction items sit on items selected for surviving that pipeline, which makes Hindi instruction accuracy optimistic relative to an unfiltered draw from the source. The first-pass rate stands as measured, and so do the construction artifacts that survived validation. Those include two math items, one Korean and one Hindi, whose translated statement did not match the English one on the automatic count of numeric quantities. They were kept rather than retranslated, because that mismatch is one the checker records as not invalidating an item.

This weakens the language axis as evidence about native non-English traffic. It weakens the instruction task type in all three languages, and the Hindi instruction items most. It weakens any per-language reading of a one-item difference, which is exactly the margin separating the per-language leaders in \Cref{sec:rule} on 98 items each. It does not weaken the identifier alignment, since one identifier denotes the same item in all three languages. That alignment is what makes a paired language comparison possible at all. It does weaken the further assumption the language axis needs from that alignment, that an aligned item is equally hard in all three renderings.

\subsection{The noise floor is itself estimated from two runs}\label{sec:limits:9}

Every reference this paper reads a difference against rests on a single pair of executions, and the weakest of them on a pair that is not a replication. The \Nflips of \Ncells model-query cells that flip, \Fliprate, is one realization of the difference between two runs. The \Ncomparable paired model-query cells would support an interval only under two conditions this design cannot check. The first is that there was no run-level shift between A and B. The second is that those positions behave as draws from one rate. Two runs cannot check the first, since a degraded period on the gateway during one of them would look the same as a higher per-position flip probability. The spread reported below is evidence against the second. The quantity with n = 2 is the run draw, not the number of positions.

The pooled rate also hides a wide spread across models. Per-model flip totals run from 0 for Gemini 3.7 Flash to 32 for DeepSeek V4 Flash. That is from 0\% to 10.88\% of a model's \Nitems model-query cells, or of \Nitemscapped for the two with an infrastructure null. The lookup holds both, 6 task$\times$language cells each. Weighting the per-model rates by the cells each model holds gives roughly 5.8\%, above the pooled figure rather than below it. That weighting is itself approximate, for two reasons. The per-model rates are whole-matrix while coding carries 73 of the \Nflips flips. And the selection rule maximized both-run accuracy, correctness in both runs. That preferentially chose each model in cells where it did not flip, and that pushes the realized exposure the other way. Which side of \Fliprate a given policy sits on is not settled by the pooled rate.

The other two references are weaker still. The \Billeddelta cost movement is one difference between two totals rather than a distribution. And \Smokeitems of \Nitems items = \Smokerate is a lower bound. It counts the items that one model scored correct under the pre-raise cap configuration and the later re-run under the raised caps did not. That model was run over the whole item set, and the pre-raise caps were 1024 by default, 2048 for instruction and 4096 for coding. A configuration change is therefore mixed into what that reference reports as execution noise.

Both directions of the paper's reading lean on these numbers. Four readings sit on the negative side. \Cref{sec:rule}'s ranking margins are one, where nine of fourteen models sit within the flip rate of the leader and seven within the per-item reference. One of those seven is on \Nitemscapped items, and the seventh is exactly at it. The three non-coding within-type items that \Cref{sec:decomp} reports as unresolved by this setup are another. \Cref{sec:lookup:2}'s reading of the unsigned-margin tie rule's own five-item loss, 5 of \Nitems items, against both accuracy references. And \Cref{sec:lookup:5}'s \LookupVsSymItems of \Nitems items = \LookupVsSymRate, the accuracy edge the adopted lookup holds over the cheaper symmetric policy, which on this reading the setup does not resolve. The comparison \Cref{sec:decomp} is built around is of the same kind and carries the same dependence. Its \ResidualArgmaxRate residual is called small by being read against both accuracy references, and a different estimate of either would move that reading directly. Which of the two would move it is not symmetric. The per-item value is the smaller of the pair, so under \Cref{sec:noise:2} it cannot resolve a difference the flip rate leaves unresolved. A per-item reference one item smaller would put the single-run residual of \Smokerate into the band where the two references disagree. The reading would then rest on the procedure's conservative branch rather than on both references agreeing. The value that decides is the flip rate, and two readings turn on it at estimates close to the one this matrix gave. Both are rates on items set against a rate on cells, which is the coarser of the two comparisons \Cref{sec:noise:2} describes. The adopted policy's \LookupVsOpusItems-item margin over the best single model, 5.78\% of items, is resolved under the \Fliprate flip rate this pair of runs produced. It would be unresolved under any estimate that reaches 5.78\% itself. \Cref{sec:rule}'s ninth-ranked model, 13 of \Nitems items = 4.42\%, is unresolved under \Fliprate and would be resolved under any estimate below 4.42\%. The first of those two is the margin the adoption rests on. No other reading in this paper is reversed by so small a change in the reference that decides it. Two readings sit on the positive side. \Cref{sec:decomp}'s statement that the three variant gaps computed without exclusions, 8.5 to 10.5 points, clear both accuracy references. And its statement that each of the three languages shows a gap of 7.14\% to 10.20\% of its 98 items that clears them too. Those checks are layered on top of the pilot's 5-percentage-point gate (\Cref{sec:decomp:1}), which reads no floor. But as floor-clearing statements they are made against a two-run reference. The unsigned-margin tie rule is not of this form. Its domain is a sign test on A and B and reads no floor value, so a different floor would not mechanically return the five items it gives up. What a different floor would bear on is the rationale for treating an unsigned margin as noise in the first place.

This weakens the paper's refusals and its floor-framed affirmatives alike, in proportion to how close each sits to the reference it is read against. The point estimates do not depend on it. The decomposition shares, the A/B billed totals, and the variant tables of \Cref{sec:decomp} and \Cref{sec:rule} are computed without reference to any floor. \LookupCorrect/\Nitems and \LookupCost are partial exceptions. Their arithmetic reads no floor, but the policy they describe was composed under a floor-motivated rule. We do not correct for the uncertainty in the floors. The procedure of \Cref{sec:noise:2} is applied as stated, with this entry as its limitation.

\section{Conclusion}\label{sec:concl}

This is a routing paper, and the number we would carry to another study is not about routing. We ran the identical configuration twice at temperature 0 and seed 42. The correct bit changed on \Nflips of \Ncells model-query cells, \Fliprate, and billed cost moved \Billedmove. Nothing was varied to produce either.

Against that movement, most of the distance from the best single model to the oracle was structure we could have written down in advance. Giving each task type its own model recovers \BetweenStable of the \GapStableAll items. Adding language recovers \LangGainArgmaxStable more. What is left, \ResidualArgmax items of \Nitems, is what a learned router would have been built to capture, and it is smaller than what a rerun moves. Seven of the \WithinStable items left inside a task type are coding.

The identity of the best single model is also not a fact about the matrix. Five models hold that title across ten scoring rules, and no top-two margin exceeds \TopTwoMargin of \Nitems items. Routing gains are usually quoted against that baseline, so a reported gain should say which rule scored it.

The pilot adopted a static task$\times$language table. It scores \LookupCorrect of \Nitems items at \LookupCost per run against Claude Opus 5's \BestStableAll at \OpusCost, ahead on both axes on an identical workload. The preregistered gate is a floor rather than an instruction: it said to build only if the distance cleared 5 percentage points of items. It passed, and we did not build one, on the strength of the decomposition above and with the results in view.

None of this shows that routers are unnecessary. What it shows is that the question worth asking first is not how complicated the router should be. It is how much of the distance is already visible in the request. Here \StructureStable of the \GapStableAll items were, and the table that took them was fixed in advance.

The scope is 14 models from one catalog snapshot, 7 task types, 3 languages, 14 items per task$\times$language cell, one gateway, two runs, and selectors fitted and evaluated on this same matrix (\Cref{sec:limits}). Correctness here is a rule-based binary check, so the matrix says nothing about work whose answers are judged rather than matched. Each task type draws all its items from one source, so task type and corpus are the same partition here.

Both of the limits that bind hardest are ours to fix. One source per task type, and a floor estimated from a single pair of runs. Two sources inside one task type would separate the axis. A third run would put an interval on the floor. Neither is expensive, and we would do both before reading a residual this small again.

\bibliographystyle{plainnat}
\bibliography{refs}

\begin{thebibliography}{8}
\providecommand{\natexlab}[1]{#1}
\providecommand{\url}[1]{\texttt{#1}}
\expandafter\ifx\csname urlstyle\endcsname\relax
  \providecommand{\doi}[1]{doi: #1}\else
  \providecommand{\doi}{doi: \begingroup \urlstyle{rm}\Url}\fi

\bibitem[Chen(2026)]{routinggapnoise2026}
Chen.
\newblock How much of the routing gap is real? decomposing the router-to-oracle
  gap into reproducible specialist advantage and single-draw label noise.
\newblock \emph{arXiv preprint arXiv:2607.03436}, 2026.
\newblock National Yang Ming Chiao Tung University and Krixvon AI.

\bibitem[Hu et~al.(2024)Hu, Bieker, Li, Jiang, Keigwin, Ranganath, Keutzer, and
  Upadhyay]{routerbench2024}
Hu, Bieker, Li, Jiang, Keigwin, Ranganath, Keutzer, and Upadhyay.
\newblock {RouterBench}: A benchmark for multi-{LLM} routing system.
\newblock \emph{arXiv preprint arXiv:2403.12031}, 2024.
\newblock ICML 2024 Workshop on Agentic Markets.

\bibitem[Huang et~al.(2025)Huang, Ling, Lin, Chen, Zhong, Wu, and
  Lin]{routereval2025}
Huang, Ling, Lin, Chen, Zhong, Wu, and Lin.
\newblock {RouterEval}: A comprehensive benchmark for routing {LLMs} to explore
  model-level scaling up in {LLMs}.
\newblock In \emph{Findings of the Association for Computational Linguistics:
  EMNLP 2025}, pages 3860--3887, 2025.
\newblock \doi{10.18653/v1/2025.findings-emnlp.208}.
\newblock Preprint arXiv:2503.10657, whose v1 author list carries an eighth
  name. Sun Yat-sen University.

\bibitem[Lee(2026)]{absencedetectiontax}
Janghoon Lee.
\newblock Repair, not improvement: Decomposing constrained decoding in
  tool-call abstention.
\newblock \emph{arXiv preprint arXiv:2608.13959}, 2026.
\newblock Posted 2026-08-14. Released with its pre-registration, data and
  harness.

\bibitem[Lee(n.d)]{companion}
Janghoon Lee.
\newblock Construction-validity companion note, n.d.
\newblock In preparation.

\bibitem[Li et~al.(2026)Li, Zhang, Guo, Wang, Tang, Zhang, Chen, Qi, Ye, Bai,
  Wang, and Hu]{llmrouterbench2026}
Li, Zhang, Guo, Wang, Tang, Zhang, Chen, Qi, Ye, Bai, Wang, and Hu.
\newblock {LLMRouterBench}: A massive benchmark and unified framework for {LLM}
  routing.
\newblock In \emph{Findings of the Association for Computational Linguistics:
  ACL 2026}, pages 37733--37754, 2026.
\newblock \doi{10.18653/v1/2026.findings-acl.1881}.
\newblock NWPU and Shanghai AI Laboratory.

\bibitem[Lu et~al.(2026)Lu, Zhang, Zhang, Yu, Wang, Chen, and
  Xing]{plateau2026}
Lu, Zhang, Zhang, Yu, Wang, Chen, and Xing.
\newblock The routing plateau: Understanding and breaking the accuracy limits
  of {LLM} routers.
\newblock \emph{arXiv preprint arXiv:2606.07587}, 2026.
\newblock Rice University and Amazon.

\bibitem[Ong et~al.(2024)Ong, Almahairi, Wu, Chiang, Wu, Gonzalez, Kadous, and
  Stoica]{routellm2024}
Ong, Almahairi, Wu, Chiang, Wu, Gonzalez, Kadous, and Stoica.
\newblock {RouteLLM}: Learning to route {LLMs} with preference data.
\newblock \emph{arXiv preprint arXiv:2406.18665}, 2024.

\end{thebibliography}

\appendix
\section{The lookup in full}\label{app:lookup}

\Cref{sec:lookup} describes the adopted policy and prices the readings of the
tie rule that produced it, but does not print the policy itself.
\Cref{tab:app-lookup} does. The \emph{basis} column is the evidence grade the
artifacts record for each cell: \emph{signed} means the selection rests on an
accuracy margin whose sign holds across both runs, and the two unsigned
grades are the ones \Cref{sec:lookup:1} counts.

\begin{table*}[t]
\centering\footnotesize
\setlength{\tabcolsep}{3.5pt}
\caption{The adopted task$\times$language lookup, all \Ntlcells\ cells.
Accuracy is both-run correct out of the \Npercell\ items in the cell; billed
cost is the mean of the two runs.}
\label{tab:app-lookup}
\begin{tabular}{@{}llllrl@{}}
\toprule
\hd{task type} & \hd{lang} & \hd{model} & \hd{correct} & \hd{mean billed} & \hd{basis} \\
\midrule
knowledge & ko & \texttt{g37f} & 13/14 & \$0.011409 & signed \\
knowledge & en & \texttt{opus} & 14/14 & \$0.059090 & signed \\
knowledge & hi & \texttt{g37f} & 13/14 & \$0.010892 & signed \\
math & ko & \texttt{dsflash} & 13/14 & \$0.001020 & unsigned (tie rule) \\
math & en & \texttt{dsflash} & 14/14 & \$0.000630 & signed \\
math & hi & \texttt{dsflash} & 14/14 & \$0.000969 & signed \\
instruction & ko & \texttt{dspro} & 12/14 & \$0.041754 & signed \\
instruction & en & \texttt{g37f} & 12/14 & \$0.032221 & signed \\
instruction & hi & \texttt{g37f} & 12/14 & \$0.038240 & signed \\
extraction & ko & \texttt{luna} & 14/14 & \$0.001471 & signed \\
extraction & en & \texttt{luna} & 14/14 & \$0.001057 & signed \\
extraction & hi & \texttt{luna} & 13/14 & \$0.001429 & unsigned (tie rule) \\
toolcall & ko & \texttt{dsflash} & 12/14 & \$0.000829 & unsigned (tie rule) \\
toolcall & en & \texttt{g37f} & 13/14 & \$0.008040 & signed \\
toolcall & hi & \texttt{g37f} & 13/14 & \$0.008287 & tie-break selection; no signed accuracy premium \\
abstention & ko & \texttt{dsflash} & 13/14 & \$0.001115 & unsigned (tie rule) \\
abstention & en & \texttt{dsflash} & 13/14 & \$0.001105 & unsigned (tie rule) \\
abstention & hi & \texttt{luna} & 14/14 & \$0.005979 & signed \\
coding & ko & \texttt{fable} & 9/14 & \$2.924815 & signed \\
coding & en & \texttt{grok43} & 9/14 & \$0.087865 & signed \\
coding & hi & \texttt{grok43} & 8/14 & \$0.094694 & unsigned (tie rule) \\
\bottomrule
\end{tabular}

\end{table*}

\section{Scoring variants and the model pool}\label{app:variants}

\Cref{sec:decomp:1} reports that all \Nvariants\ scoring variants keep the
oracle above the best single model, and \Cref{sec:rule} reports that the
identity of the best single model moves across them.
\Cref{tab:app-variants} gives each variant.

\begin{table*}[t]
\centering\footnotesize
\setlength{\tabcolsep}{3.5pt}
\caption{The \Nvariants\ scoring variants. Distances are in percentage
points of items. Where a variant excludes items per model, the denominators
differ between rows and the distances are differences between ratios.}
\label{tab:app-variants}
\begin{tabular}{@{}lllll@{}}
\toprule
\hd{scoring variant} & \hd{best single} & \hd{best} & \hd{oracle} & \hd{distance (pp)} \\
\midrule
single-run A, no exclusion & \texttt{fable} = \texttt{opus} & 249/294 & 280/294 & 10.5 \\
both-run, no exclusion & \texttt{opus} & 245/294 & 274/294 & 9.9 \\
single-run A, cap-hits excluded & \texttt{qwen} & 248/282 & 280/294 & 7.3 \\
both-run, cap-hits excluded & \texttt{qwen} & 238/280 & 274/294 & 8.2 \\
single-run B, no exclusion & \texttt{opus} & 252/294 & 277/294 & 8.5 \\
single-run A, parse failures excluded & \texttt{fable} & 249/290 & 280/294 & 9.4 \\
both-run, parse failures excluded & \texttt{g37f} & 244/288 & 274/294 & 8.5 \\
\bottomrule
\end{tabular}

\end{table*}

\Cref{tab:app-pool} gives the \Nmodels\ models with the three quantities the
body uses them for: their both-run score, their billed cost on each run, and
how many of their \Nitems\ model-query cells changed their correct bit
between the two runs.

\begin{center}
\footnotesize
\setlength{\tabcolsep}{3.5pt}
\captionof{table}{The \Nmodels-model pool. Short names are the ones
\Cref{sec:setup} introduces. Flips are of \Nitems\ model-query cells per
model, and sum to \Nflips.}
\label{tab:app-pool}
\begin{tabular}{@{}llrrr@{}}
\toprule
\hd{model} & \hd{both-run correct} & \hd{billed A} & \hd{billed B} & \hd{flips} \\
\midrule
\texttt{opus} & 245/294 & \$8.311260 & \$7.066535 & 11 \\
\texttt{g37f} & 244/294 & \$0.468923 & \$0.469771 & 0 \\
\texttt{g31p} & 243/294 & \$7.109740 & \$6.938848 & 4 \\
\texttt{sol} & 239/293 & \$1.293286 & \$1.283346 & 7 \\
\texttt{fable} & 238/294 & \$14.733090 & \$14.189540 & 17 \\
\texttt{qwen} & 238/294 & \$1.071350 & \$1.075630 & 15 \\
\texttt{kimi} & 236/294 & \$7.689702 & \$7.746045 & 15 \\
\texttt{terra} & 234/294 & \$1.512234 & \$1.534734 & 7 \\
\texttt{sonnet} & 232/294 & \$4.351772 & \$4.460802 & 28 \\
\texttt{dspro} & 227/294 & \$0.923519 & \$0.740010 & 18 \\
\texttt{luna} & 223/293 & \$0.174658 & \$0.169133 & 17 \\
\texttt{grok46} & 220/294 & \$3.156662 & \$3.067956 & 23 \\
\texttt{grok43} & 218/294 & \$0.686402 & \$0.709431 & 27 \\
\texttt{dsflash} & 210/294 & \$0.074890 & \$0.140151 & 32 \\
\bottomrule
\end{tabular}

\end{center}

\section{Timestamps, hashes and identifiers}\label{app:pins}

The body reports the order in which decisions were taken, because a reader
does not need our clock to follow the argument. The clock is here, so that
each decision can be matched to the record it was taken in. All times are UTC
on 2026-08-23 unless another date is given.

\begin{center}
\footnotesize
\setlength{\tabcolsep}{4pt}
\captionof{table}{Decision pins. Each row is a decision taken at the time
given and recorded in the project's decision log, and the passage it
governs. A record may carry a later header time than a decision it quotes,
so the log is searched by time rather than by filename.}
\label{tab:pins}
\begin{tabular}{@{}lll@{}}
\toprule
\hd{Time} & \hd{Decision} & \hd{Governs} \\
\midrule
02:52Z & pilot closed on a static lookup & \Cref{sec:lookup} \\
03:33Z & three coding cells held        & \Cref{sec:lookup} \\
04:01Z & pre-rule policy frozen         & \Cref{sec:lookup:5} \\
04:44Z & tie rule adopted, domain fixed  & \Cref{sec:lookup:1} \\
17:28Z & resolution procedure stated    & \Cref{sec:noise:2} \\
\bottomrule
\end{tabular}
\end{center}

The aggregation pin is 2026-08-21T13:22Z. Prices are USD per million tokens
from the 2026-08-21 catalog snapshot. Model identifiers were verified by
live call on 2026-08-22. The LiveCodeBench window is 2025-01-01 to
2025-04-06. The frozen item ledger is \Nledger\ unique item identifiers,
fixed under SHA-256 \texttt{\Ledgerhash\dots}. Neither the ledger nor the
label matrix accompanies this version, and \Cref{sec:setup:3} says what that
hash does and does not let a reader do.

The body abbreviates every hash it prints, so that its rows and sentences
stay readable in a column. \Cref{tab:hashes} gives each one in full.

\begin{center}
\footnotesize
\captionof{table}{Abbreviated hashes in full. The left column is the form the
body prints; the right is the value to pin against. Six rows are the source
dataset revisions of \Cref{sec:setup:2}. The seventh, the longest, is the
frozen item ledger of \Cref{sec:setup:3}, which is not a dataset revision.}
\label{tab:hashes}
\begin{tabular}{@{}l>{\ttfamily\raggedright\arraybackslash}p{0.56\columnwidth}@{}}
\toprule
\hd{In the body} & \hd{Full revision} \\
\midrule
\texttt{0fe84c39}\dots & 0fe84c39\allowbreak 12ea0c4d\allowbreak 4a780370\allowbreak 83943e8f\allowbreak 0c4dd505 \\
\texttt{6ea57973}\dots & 6ea57973\allowbreak c7a6097f\allowbreak d7c59156\allowbreak 98c54c17\allowbreak c5b1b6c8 \\
\texttt{740312ad}\dots & 740312ad\allowbreak d88f7819\allowbreak 78c06588\allowbreak 06c59bc2\allowbreak 815b9866 \\
\texttt{8e6106a6}\dots & 8e6106a6\allowbreak c6ce1c50\allowbreak 27e66cc3\allowbreak 38143cf9\allowbreak 97b2aa09 \\
\texttt{903af87e}\dots & 903af87e\allowbreak dfaaaa20\allowbreak 5bb083ab\allowbreak 8de7eab4\allowbreak 9ab1174d\allowbreak 85e0550e\allowbreak b1e26e1b\allowbreak e064c992 \\
\texttt{966cd895}\dots & 966cd895\allowbreak 45d6b6ac\allowbreak fd7638bc\allowbreak 708b9826\allowbreak 1ca58e84 \\
\texttt{b2f13d42}\dots & b2f13d42\allowbreak 6afe3be8\allowbreak d69a7e73\allowbreak 9b36724d\allowbreak b8b66bbc \\
\bottomrule
\end{tabular}

\end{center}

\end{document}